\pdfoutput=1
\documentclass[11pt]{article}

\usepackage[]{acl}

\usepackage{times}
\usepackage{latexsym}

\usepackage[T1]{fontenc}
\usepackage[utf8]{inputenc}

\usepackage{microtype}

\usepackage{inconsolata}

\usepackage{graphicx} 
\usepackage{tabularx}
\usepackage{multirow}
\usepackage{booktabs}
\usepackage{multirow}
\usepackage{xspace}
\newcommand{\F}{F$_1$\xspace}

\usepackage{xcolor}

\usepackage{paralist} % and \usepackage{enumitem} don't interact well

\usepackage{amsmath}
\usepackage{stackengine}
\usepackage{xcolor}

\usepackage{caption}
\usepackage{subcaption}

\renewcommand{\paragraph}[1]{\par\noindent\textbf{#1}}

\title{Understanding Fine-grained Distortions in Reports of Scientific Findings}

\author{Amelie W\"uhrl$^1$, Dustin Wright$^2$, Roman Klinger$^3$ \and Isabelle Augenstein$^2$ \\
  $^1$University of Stuttgart, Germany, 
  $^2$University of Copenhagen, Denmark,
  $^3$University of Bamberg, Germany \\
  \texttt{amelie.wuehrl@ims.uni-stuttgart.de}\\
  \texttt{\{dw, augenstein\}@di.ku.dk}\\
  \texttt{roman.klinger@uni-bamberg.de}\\
}

\begin{document}
\maketitle
\begin{abstract}

Distorted science communication harms individuals and society as it can lead to unhealthy behavior change and decrease trust in scientific institutions. Given the rapidly increasing volume of science communication in recent years, a fine-grained understanding of \emph{how} findings from scientific publications are reported to the general public, and methods to detect distortions from the original work automatically, are crucial. Prior work focused on individual aspects of distortions or worked with unpaired data. In this work, we make three foundational contributions towards addressing this problem: (1) annotating 1,600 instances of scientific findings from academic papers paired with corresponding findings as reported in news articles and tweets wrt. four characteristics: causality, certainty, generality and sensationalism; (2) establishing baselines for automatically detecting these characteristics; and (3) analyzing the prevalence of changes in these characteristics in both human-annotated and large-scale unlabeled data. Our results show that scientific findings frequently undergo subtle distortions when reported. Tweets distort findings more often than science news reports. Detecting fine-grained distortions automatically poses a challenging task. In our experiments, fine-tuned task-specific models consistently outperform few-shot LLM prompting.

\end{abstract}

\section{Introduction}
Lay people, i.e., non-experts with limited experience or knowledge of a specific domain, rely on effective science communication to learn about scientific research. In order to make scientific information understandable to a lay audience, science communicators must first simplify the highly technical language of science~\cite{salita2015writing}. In doing so, authors may knowingly or unknowingly distort the information conveyed by the original scientific publication to achieve specific rhetorical goals~\cite{ransohoff2001sensationalism,dempster2022scientific,tichenor1970mass,sumner2014association,bratton2019association}. For example, simplifying findings for a non-expert audience requires balancing accuracy, accessibility and comprehensibility \citep{kuehne-olden-2015} which can lead to information being omitted purposefully. At the same time, the way that science is communicated to the public is crucial as it influences people's behavior and trust in science~\cite{kuru2021effects,fischhoff2012communicating,hart2016impact}. 

\begin{figure}
    \includegraphics[scale=.7]{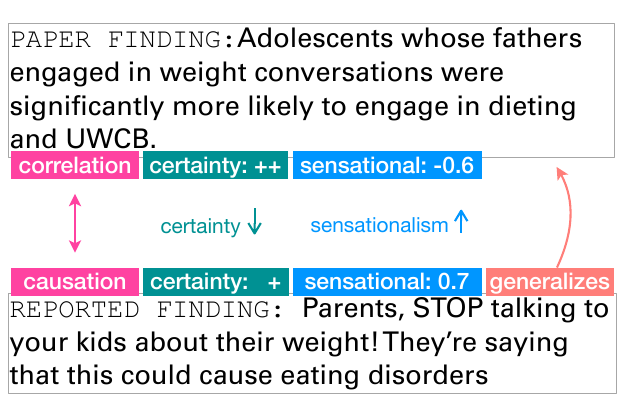}
    \caption{Pair of scientific finding and reported finding with fine-grained labels of distortions.}
    \label{fig:example-changes}
\end{figure}

Consider Fig.~\ref{fig:example-changes}. The paper finding describes a correlation between ``weight conversations'' and ``dieting and UWCB''\footnote{UWBC stands for unhealthy weight control behaviors.} while clearly stating to whom the finding applies. In the reported finding those constraints are omitted, it generalizes from ``fathers'' to all ``parents'' and ``adolescents'' to ``kids''. Further, the reported finding states a causal relationship between ``talking to kids about their weight'' and ``eating disorders''. Finally, the reported finding expresses high certainty in the correlational relationship found, while the reported finding speculates that a causal relationship ``could'' exist, thus communicating lower certainty. While subtle, the reported finding presents a different picture than that of the original paper, highlighting the need for a fine-grained comparison between paper and reported findings.

Previous work has been limited to either a subset of the distortions studied in this work~\cite{yu-etal-2019,pei-jurgens-2021,wright-augenstein-2021} or detecting general information change without fine-grained labels~\cite{wright-et-al-2022}. We improve on this by making three core contributions: We collect an expert-annotated dataset of 1,600 scientific findings paired across scientific papers and their reported findings in news media and Twitter (\textbf{C1}).\footnote{The data will be publicly available upon acceptance.} The data is annotated with fine-grained distortion labels, i.e., causal claim strength, level of certainty, level of generality, and sensationalism. Using this labeled data, we train and analyze the performance of benchmark models on the task of fine-grained distortion detection \textbf{(C2)}. Finally, using the data from \textbf{C1} and models from \textbf{C2}, we perform a large scale analysis of the prevalence and types of distortions present in both our expert labeled data and a large-scale automatically labeled dataset of 1,655,570 paper findings, 422,626 news findings and 356,275 tweets \textbf{(C3)}.
We answer the following research questions:
\begin{compactitem}
\item[\textbf{RQ1}] How are scientific findings changed when reported to lay audiences? 
    \item[\textbf{RQ2}] How reliably can we detect distortions automatically? 
\end{compactitem}

For RQ1, we find that scientific findings undergo fine-grained changes when they are reported even when their overall content is well-aligned. This is consistent across scientific disciplines. We find that 54 \% of findings are reported with a changed causal relation and 60 \% of findings are reported with a changed level of certainty. In 49 \% of the paired findings, the reported finding is more general than the paper finding, and reported findings are typically more sensational compared to paper findings. Across all change dimensions, findings reported in tweets are more susceptible to mis-reporting compared to reports in science news. 
With respect to RQ2, we find that detecting fine-grained distortions automatically poses a challenging task. In our experiments, fine-tuning task-specific models consistently outperform few-shot LLM prompting. Our best models achieve macro \F scores of 0.58, 0.56, 0.57 and Pearson correlation of 0.61 for predicting causality, certainty, generalizing and sensationalism, respectively.  

\section{Related Work}
\paragraph{Science Communication.}
Science communication is a relatively nascent area of exploration in NLP. The main problems which have been worked on  relate to information change in news articles and social media about scientific papers~\cite{wright-et-al-2022,wright-augenstein-2021,pei-jurgens-2021,yu-etal-2020-measuring}, understanding discourse strategies in scientific press releases~\cite{august-et-al-2020}, information loss in medical summaries~\citep{trienes2024}, tasks related to scientific peer review~\cite{DBLP:journals/coling/KuznetsovBEG22}, tasks related to scholarly document understanding~\cite{DBLP:conf/acl/WrightA21,DBLP:conf/emnlp/BeltagyLC19} and scientific fact checking~\cite{DBLP:conf/emnlp/WaddenLLWZCH20,DBLP:conf/lrec/MohrWK22}.
This work is most closely related to those studying information change in science communication, particularly the works of~\cite{wright-et-al-2022} on general information change and \cite{wright-augenstein-2021} and \cite{pei-jurgens-2021} on exaggeration and certainty, respectively. These works are limited in a few key aspects, which we address. First, they are concerned with either single, narrow aspects of information change or overarching broad notions of change, missing important types of distortions such as generalizing and sensationalizing results. Additionally, the existing labeled data for exaggeration is limited in size, and the labeled data for certainty are unpaired. We improve on this by augmenting the matched findings in the dataset from ~\citet{wright-et-al-2022} with four specific distortions that are prevalent in science communication: exaggerating causal claim strength, changing the level of certainty, generalizing results, and sensationalizing results.

\paragraph{Misinformation.}
Inaccurate reports of scientific findings is a form of mis-information. Misinformation detection and fact-checking are established tasks in NLP, both for the general domain and for scientific claims~\citep{guo-etal-2022-survey,vladika-matthes-2023-scientific}. Scientific fact-checking verifies scientific claims against evidence sources. It is related to our task as it is compares the truthfulness of a statement against a reference document. Technically, a reported finding constitutes a claim about the original which connects our task to claim detection and argument mining~\citep{lawrence-reed-2019-argument,boland2022}. However, compared to both these related tasks, this work requires a more nuanced view of detailed characteristics of the overarching claim. 

\section{Dimensions of Information Changes}
We consider four dimensions found to be notable in the science of science literature~\cite{sumner2014association,bratton2019association,fischhoff2012communicating,ransohoff2001sensationalism} that characterize scientific findings and may undergo change when reported: \textit{Causality}, i.e., the type of causal relation (or its absence) described in finding; \textit{Certainty}, i.e., the level of confidence or certainty that is expressed wrt. a finding; \textit{Generality}, i.e., the level of generalization or specificity of a finding compared to its reporting; \textit{Sensationalism}, i.e., the extent to which a finding is presented in a way to elicit an emotional reaction by using urgent or exaggerated language and descriptions.
The divergence between those dimensions in the paper and the reported findings allows us to estimate how accurate a reporting is and which properties may be distorted. More specifically, we consider a reported finding to be mis-reported if the label for a given characteristic changes from the paper finding to the corresponding reported finding.
We build a dataset of 1,600 findings from four scientific disciplines (medicine, psychology, biology and computer science). Findings are paired between scientific paper and news and scientific paper and tweet, giving 800 pairs total. In an annotation conducted with crowdworkers, we label each instance/pair with regards to the change type. Annotators rate certainty and sensationalism levels, identify causality relations and check for generalizations in both versions of the finding. Fig.~\ref{fig:example-changes} shows an example.

\subsection{Change Dimensions}
(1) \textbf{Causality} describes a cause--effect relationship between two things, variables, agents etc. Correlation describes relationships where two actions relate to each other, but one is not necessarily the effect or outcome of the other. %Linguistic cues for causality are words like `cause' or `decrease' that describe how X directly causes outcome Y. Cues for correlation include descriptions such as `associated with', `connection', or `linked to'.
(2) \textbf{Certainty} in science communication can be expressed with respect to various aspects \citep{pei-jurgens-2021-measuring}. (Un)certainty exists towards specific numbers/quantities (`approximately 50 \%'), the extent to which a finding applies (`mainly observed for') or the probability that something applies, occurs or is associated (`possibly associated with'). % Linguistic cues for this concept include hedging words e.g., `seem', `tend', `may', `potentially' or `suggest'.
(3) \textbf{Generalizations} claim something is always true, even if it is only valid in certain instances or occasionally. For example, a reporting that generalizes a finding about diabetes type 2 in seniors to all people with diabetes or a generalization from a specified set of medical conditions (`reduced risk of stroke and diabetes') to a general statement (`has health benefits').
(4) Sensational text intends to spark the interest of a reader, make them curious or elicit an emotional reaction. Cues for \textbf{sensationalism} can be urgent, exaggerated language %: `life-changing', `unparalleled performance', or `revolutionary'. 
or conveyed through the use of informal or colloquial language.%, e.g., expressions like `amps up the efficiency' or `They ran some solid experiments to back this up'.

\subsection{Dataset Construction}
\subsubsection{Source Data}
To investigate \emph{how} science communication changes scientific findings, we require reports of findings matched with their original counterpart.
Therefore, we build on \textsc{Spiced} \citep{wright-et-al-2022} which provides text pairs of scientific findings and associated reports of the finding from news articles or Twitter. Each pair is scored with an \textit{information matching score (IMS)} which indicates how similar the content of the two texts are. It ranges from \textit{5 -- completely the same} to \textit{1 -- completely different}. 
We sample instances with a high information matching score (IMS $> 4$) and filter out instances with a high IMS that the \textsc{Spiced} dataset marks as \textit{easy} cases\footnote{\textsc{Spiced} marks instances as \textit{easy} if the reported finding is almost identical to the paper finding.}.

The filtering provides us with a total of 837 paired findings across four scientific disciplines: biology (185), computer science (168), medicine (227) and psychology (257). The reported findings stem from science news (515) and Twitter (322).

\subsubsection{Annotation Tasks}
Considering the substantial differences in change type concepts, we design the data collection as four separate annotation tasks to enable annotators to focus on one concept at a time.\footnote{In a set of pilot studies, we experiment with tasking annotators to label all change types for an instance, instead of focusing on one concept per study. However, inter-annotator agreement in this setup was very low, presumably because of the difficulty the task and the substantial cognitive complexity it takes to understand and switch between multiple concepts during annotation.} This also allows us to operationalize each task independently, which is important to find the optimal annotation method for each task without burdening the annotators with strenuous context switches. We describe the tasks and settings in the following.
\paragraph{Causality.}
Given a finding, annotators are tasked to identify which type of causal relationship is described. In a classification setting, annotators decide between \textit{No relation stated}, meaning no causal or correlational relation is stated, \textit{Correlation}, \textit{Causation} and \textit{Explicitly states: no relation}, meaning the finding states the absence of a relation.
\paragraph{Certainty.}
Given a finding, annotators rate the level of certainty with which a finding is being described. This task uses a 4-point rating scale ranging from \textit{Uncertain} to \textit{Certain} with the nuances \textit{somewhat uncertain} and \textit{somewhat certain} in between.
\paragraph{Generalization.} Given a paired finding, annotators identify which finding is more general, i.e., the \textit{reported finding}, or the \textit{paper finding}. If they are equally specific/general, annotators can label them as expressing the \textit{same level of generality}.
\paragraph{Sensationalism.}
Annotators are presented with sets of four findings at a time. In a best-worst-scaling setup \citep{kiritchenko-mohammad-2017-best}, they identify which of the four findings is the \textit{most} and \textit{least sensational}.

\subsubsection{Annotation Environment}
We use \textsc{Potato} \citep{pei2022} as our annotation environment and recruit crowdworkers using \textsc{Prolific}.\footnote{\url{www.prolific.co}} To ensure subject expertise, participants must have at least an undergraduate degree in the respective scientific field or a closely related subject (refer to Appendix \ref{appendix:annotation} for details.) For the change types \emph{causality}, \emph{certainty}, and \emph{generalization}, every annotator works on 12 instances. For \emph{sensationalism}, participants work on 10 instances, i.e., quad-tuples.\footnote{Quad-tuples contain a mixture of paper findings and reported findings. To avoid biasing the annotators, we do not make it transparent to annotators which source the individual finding originated from.} We provide a detailed description of the annotation setup in Appendix~\ref{appendix:annotation} and screenshots in the supplementary material.

\subsubsection{Label Aggregation}
For \textit{causality}, \textit{certainty} and \textit{generalization}, we aggregate the final labels using \textsc{MACE}~\citep{hovy-etal-2013-learning}, a Bayesian model which learns a distribution over labels that takes into account annotator competence.
To obtain real-valued scores from best-worst annotation, we calculate the percentage of times an instance was chosen as most sensational minus the percentage of times the term was chosen as least sensational~\citep{kiritchenko-mohammad-2017-best}. The score ranges between $-$1 and 1.

\subsection{Analysis}
\subsubsection{Evaluation Metrics}
\label{eval-metrics}
We evaluate the results of the annotation studies using the following metrics: Average pairwise inter-annotator \F (ia\F), i.e.,  treating one annotator’s labels as  gold annotations and consider the other annotator’s labels as predictions~\citep{hripcsak_agreement_2005}; average pairwise Cohen's $\kappa$; average\footnote{We average correlations by transforming each correlation coefficient using Fisher's Z, calculating the average of the transformed values, and back-transforming the value.} pairwise Spearman's correlation $\rho$; split-half reliably to evaluate best-worst scaling tasks~\citep{kiritchenko-mohammad-2017-best} for which all annotations for an instance are split into half. For each set, the best-worst scaling score is calculated independently. We report the correlation (Spearman and Pearson) between the sets of scores\footnote{We use the best-worst-scaling scripts available at \url{https://saifmohammad.com/WebPages/BestWorst.html}}.

\subsubsection{Agreement}
We report agreement metrics for all tasks in Table~\ref{tab:final-data_agreement}. For \textit{causality}, we observe an inter-annotator \F of 0.38. The average pairwise $\kappa$ agreement is 0.21 indicating fair agreement \citep{McHugh2012}. For \textit{certainty}, the average correlation ($\rho$) between the certainty ratings is 0.44. In the \textit{generalization} task, we observe an inter-annotator \F of 0.42 and a $\kappa$ of 0.20. We report split-half reliability for the \textit{sensationalism} task and observe an average $\rho$ of 0.44 indicating a positive correlation between the sets of scores.

We acknowledge that the agreement scores are in parts relatively low. We presume that this reflects the difficulty of the tasks as judging scientific findings is not trivial and sensationalism and certainty are to a certain extend subjective. Note that is in line with agreement scores reported for similar tasks (e.g., classifying writing strategies for science communication \citep{august-et-al-2020}).

 \begin{table}[ht]
        \centering\small
        \begin{tabularx}{\linewidth}{XXXXXXXXXX}
            \toprule
            & \multicolumn{2}{c}{causality}& 
            \multicolumn{3}{c}{certainty}&
            \multicolumn{2}{c}{general.}&
            \multicolumn{2}{c}{sensation.}\\

        \cmidrule(lr){2-3} \cmidrule(lr){4-6} \cmidrule(lr){7-8} \cmidrule(l){9-10}
            
            disc.
            % causal
            &ia\F  & $\kappa$
            % certainty
            & ia\F  & $\kappa$ & $\rho$
            % general
            & ia\F  & $\kappa$ 
            % sensational
            & $\rho$  & $r$ \\

            \cmidrule(r){1-1}\cmidrule(lr){2-2}\cmidrule(lr){3-3}\cmidrule(lr)
            {4-4}\cmidrule(lr){5-5}\cmidrule(lr){6-6} \cmidrule(lr){7-7} \cmidrule(lr){8-8} \cmidrule(lr){9-9}\cmidrule(l){10-10} 
            
            bio &
            % causal
            .40 & .22 
            % certainty
            & .37& .22&.48 
            % general
            & .43& .24
            % sensational
            &.48& .48
            \\

            cs &
            % causal
            .36&.20
            % certainty
            &.31 & .15& .42
            % general
            & .41&.17
            % sensational
            &.47&.48 
            \\

            med &
            % causal
            .38&.23 
            % certainty
            &.36&.19&.48
            % general
            & .44&.23
            % sensational
            &.41&.44 
            \\

            psy &
            % causal
            .36&.20
            % certainty
            &.37&.22&.38
            % general
            & .40&.16
            % sensational
            &.38&.39 
            \\

            % averages
            \cmidrule(lr){2-3} \cmidrule(lr){4-6} \cmidrule(lr){7-8} \cmidrule(l){9-10}

            avg. &
            % causal
            .38 &.21
            % certainty
            &.35&.20&.44
            % general
            & .42&.20
            % sensational
            & .44&.45 \\
            \bottomrule
        \end{tabularx}
        \caption{Inter-annotator \F (ia\F), average pairwise Cohen's $\kappa$, Spearman's correlation ($\rho$) across tasks and disciplines. For sensationalism, $\rho$ and $r$ are the correlations from the split-half reliability evaluation.}
        \label{tab:final-data_agreement}
    \end{table}

\subsubsection{Results: How are scientific findings changed when they are reported to lay audiences? (RQ1)}
\begin{figure*}
     \centering
     \begin{subfigure}[b]{0.25\textwidth}
         \centering
         
         \includegraphics[scale=0.27]{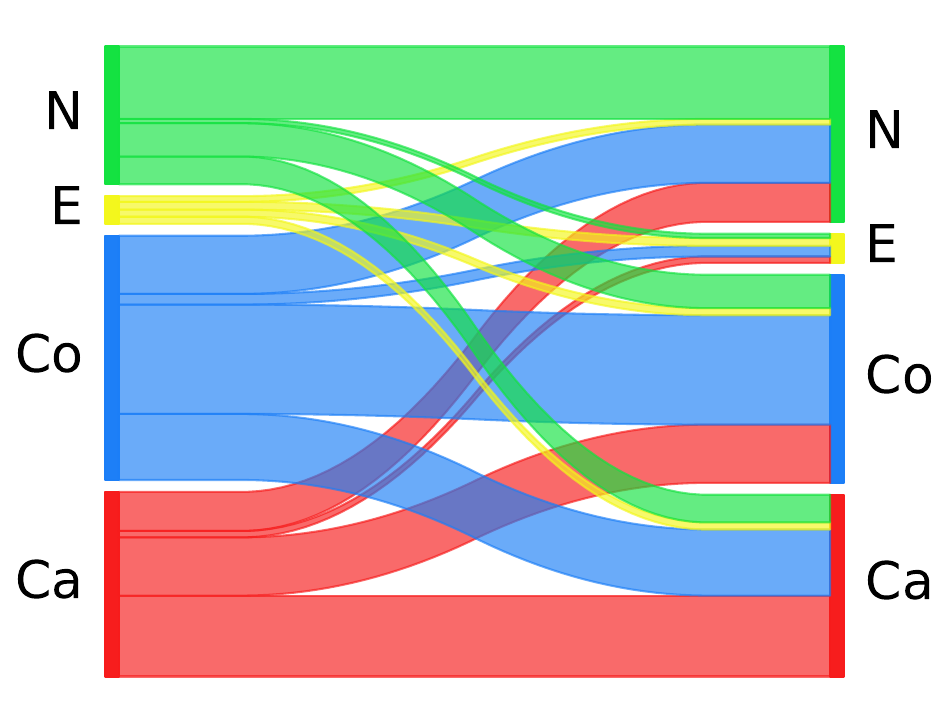}
         
         \caption{Causality. N: No rel. mentioned, E: Expl. states: no rel., Co: Correlation, Ca: Causation.}
         \label{fig:causal-changes}
         
     \end{subfigure}
     \hfill
     \begin{subfigure}[b]{0.25\textwidth}
         \centering
         
         \includegraphics[scale=0.27]{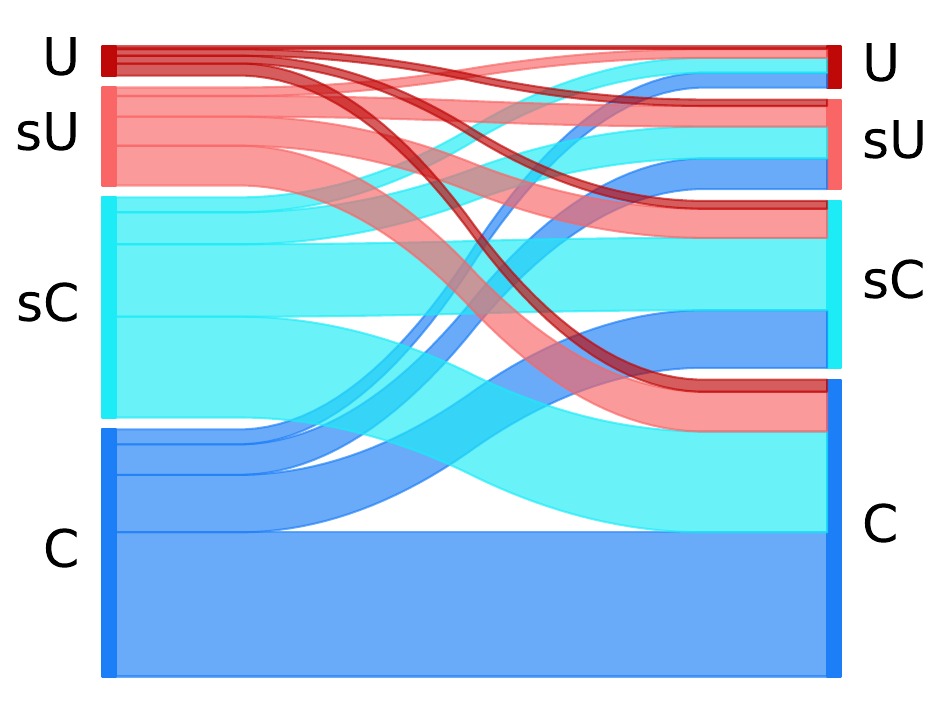}
         \caption{Certainty. (s)U: (somewhat) Uncertain, (s)C: (somewhat) Certain.}
         \label{fig:certainty-changes}
     \end{subfigure}
     \hfill
     \begin{subfigure}[b]{0.2\textwidth}
         \centering
         
         \includegraphics[scale=0.24]{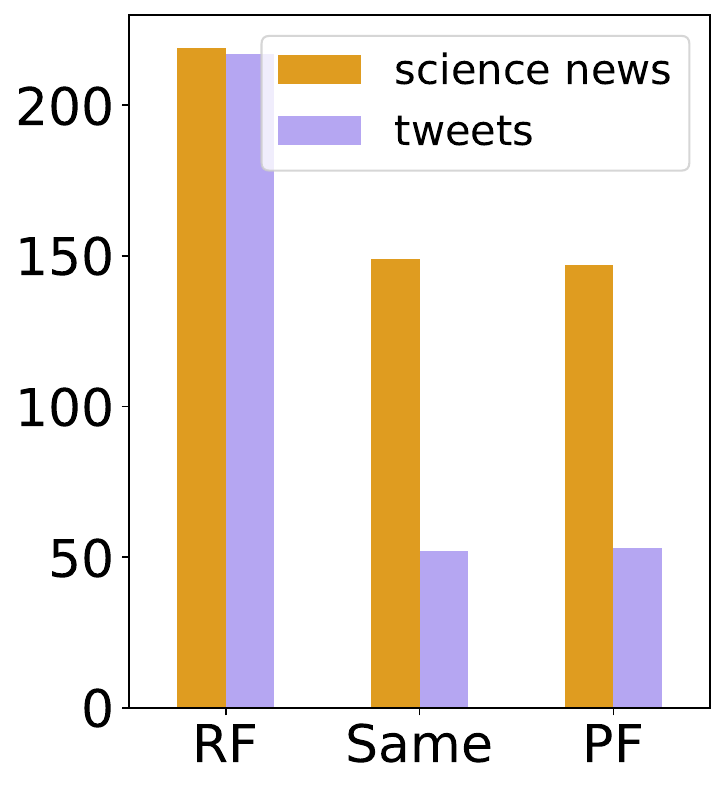}
         \caption{Which finding is more general? Rep./Paper Finding (R/PF), same level.}
         \label{fig:generalization-changes}
     \end{subfigure}
     \hfill
     \begin{subfigure}[b]{0.25\textwidth}
         \centering
         
         \includegraphics[scale=0.24]{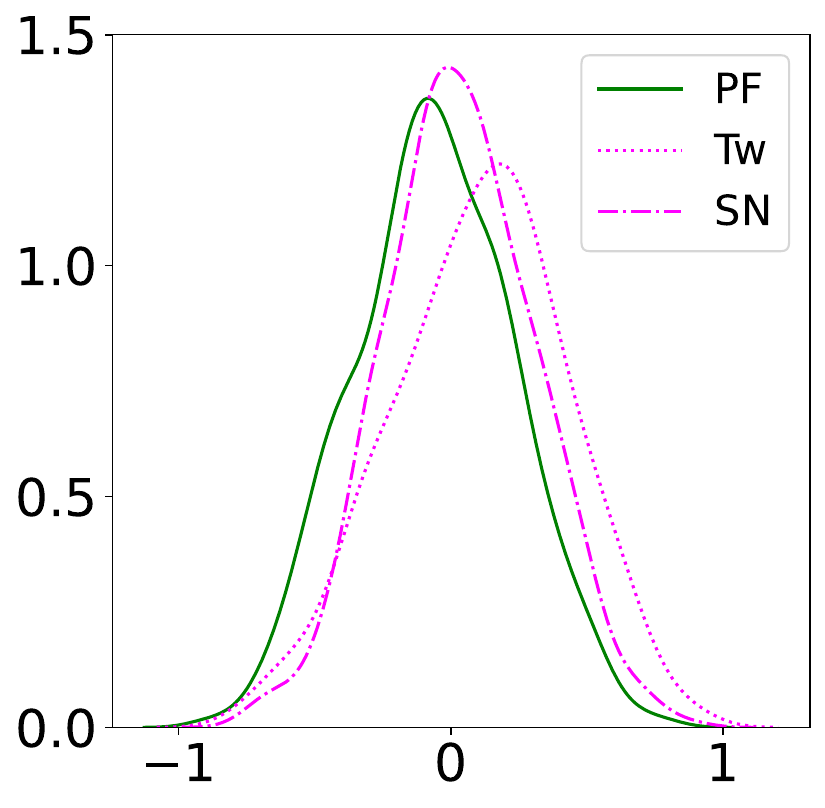}
         \caption{Sensationalism scores (x-ax.). Paper finding (PF), tweets (Tw), science news (SN).}
         \label{fig:sensational-changes}
     \end{subfigure}
     
        \caption{Sankey diagrams visualize changes in causality and certainty. The left side of the charts depict the labels for the paper finding, the right side depicts the reported finding. Bar plot visualizes the distribution of generalization labels. Density plot visualizes the difference in sensationalism scores across reporting source.}
        \label{fig:sankey-changes-visual}
\end{figure*}

We want to understand if and how scientific findings papers are distorted when they are reported to lay audiences. To this end,  we compare for each paired finding how the label for a particular dimension (causality, certainty, generalization, sensationalism ) changes going from the paper finding to the reported finding.  Fig.~\ref{fig:sankey-changes-visual} visualizes the results. Figures~\ref{fig:causal-changes} and \ref{fig:certainty-changes} present distortions in Sankey diagrams, with the left side of each chart depicting the label for the paper finding and the right side depicts the label for the reported finding. Each label is represented by a set of strands going from left to right. Figure~\ref{fig:generalization-changes} plots the distribution of labels for generalization in a bar plot. For each label we plot the number of instances separated by communication outlet. Fig. \ref{fig:sensational-changes} visualizes the sensationalism scores for the paper finding and the scores for the reported findings (tweets and science news) in a density plot. 

% causality
Fig.~\ref{fig:causal-changes} shows changes in causality. Within each relation type (causation, correlation etc.) for the paper findings, we see that the strongest Sankey strand typically leads to its same-label-counterpart on the right side (e.g., \textit{Correlation} to \textit{Correlation}). However, while these strands tend to be the strongest, the sum of the other strands originating from each group, is equally substantial. In fact, overall only 45.5\% of paired findings convey the same relation in the paper and reported finding.

% certainty
Fig.~\ref{fig:certainty-changes} shows changes in certainty. Overall we observe that both paper and reported findings typically describe the finding with relative certainty. The most frequent distortion is turning a \textit{somewhat certain} paper finding to a \textit{certain} reported finding. In general, we observe that transitions to neighboring labels on the certainty scale are typically more frequent than changes to certainty levels further away on the scale. Collapsing the \textit{somewhat} \textit{(un)certain} findings, we find that 15\% of paper findings labeled as \textit{certain} are reported in an \textit{uncertain} manner, while 13\% of paper findings labeled as \textit{uncertain} are reported in a \textit{certain} manner.

% generalization
We visualize changes in generality in Fig.~\ref{fig:generalization-changes}. Overall we observe that in the majority of cases reported findings are more general compared to the paper findings. This means findings typically start out being specific and become more general in the reporting. The opposite occurs less frequently. 

% sensatioalism
The density plots in Fig.~\ref{fig:sensational-changes} show the distribution of sensationalism scores across paper and reported findings. Reported findings are separated by communication outlet (science news, tweets). The score (x-axis) ranges from -1 (least sensational) to 1 (most sensational). We see that the majority of findings have a sensationalism score around 0. Notably, the  distribution of the reported finding scores is offset more toward $+$1, indicating that the reportings are typically more sensational compared to the paper findings.

\paragraph{Changes across communication outlets.} We want to understand if the communication outlet (science news, Twitter) impacts the types of distortions we observe.
For this, we identify four distortions which are potentially most harmful for the news consumer. Critical distortions are: \begin{compactitem}
    \item[\textbf{caus$\uparrow$}] Increase in causal claim strength: \textit{Correlation} or \textit{Explicitly States: no relation} in paper finding to \textit{Causation} in reported finding.
    \item[\textbf{gen$\uparrow$}] Increase in generality: the reported finding being \textit{more general} than the paper finding.
    \item[\textbf{cert$\uparrow$}] Increase in certainty.
    \item[\textbf{sens$\uparrow$}] Increase in sensationalism scores between from paper to reported finding $>$ 1sd.
\end{compactitem}

Table~\ref{tab:change-types-comm-outlet} shows the percentage of finding pairs affected by critical distortions across the two communication outlets, i.e., Twitter and science news. For the categories \textit{Increase in causal claim strength} (caus$\uparrow$) and  \textit{Increase in certainty} cert$\uparrow$, there are no major differences between science news and tweets. For \textit{Increase in generality} (gen$\uparrow$) however, reports in tweets are distorted substantially more frequently than reported findings in science news. Similarly, findings reported in tweets more frequently sensationalize paper findings compared to reported findings from sciences news.

\begin{table}
    \centering \small
    \begin{tabular}{lrr|rrrr}
        \toprule
         type&news  &tweets & bio & cs & med & psy  \\
         
        % causality
        \cmidrule(r){1-1}\cmidrule(lr){2-2}\cmidrule(lr){3-3} \cmidrule(lr){4-4} \cmidrule(lr){5-5} \cmidrule(lr){6-6} \cmidrule(lr){7-7} 
        caus$\uparrow$& 14.2 & 12.4& 11.9 & 18.5 & 14.1 & 10.9 \\
        % certainty
        \cmidrule(r){1-1}\cmidrule(lr){2-2}\cmidrule(lr){3-3} \cmidrule(lr){4-4} \cmidrule(lr){5-5} \cmidrule(lr){6-6} \cmidrule(lr){7-7} 
        
        cert$\uparrow$ & 33.0&32.9 &31.9 & 43.5 & 29.1 & 30.4\\

        % general.
        \cmidrule(r){1-1}\cmidrule(lr){2-2}\cmidrule(lr){3-3} \cmidrule(lr){4-4} \cmidrule(lr){5-5} \cmidrule(lr){6-6} \cmidrule(lr){7-7} 
        
        gen$\uparrow$ & 42.5& 67.4&53.5 & 42.9 & 50.2 & 58.8\\
        gen$\downarrow$ & 28.5& 16.5 &23.2 & 30.4 & 25.6 & 18.7\\

        % sensational.
        \cmidrule(r){1-1}\cmidrule(lr){2-2}\cmidrule(lr){3-3} \cmidrule(lr){4-4} \cmidrule(lr){5-5} \cmidrule(lr){6-6} \cmidrule(lr){7-7} 
        sens$\uparrow$ & 39.6 & 50.6 &43.8 & 39.9 & 40.1 & 49.8 \\
    
        \bottomrule
    \end{tabular}
    \caption{Percentage of finding pairs affected by critical distortions across communication outlets (science news, tweets) and scientific disciplines (biology, computer science, medicine, psychology).}
    \label{tab:change-types-comm-outlet}
\end{table}

\paragraph{Changes across scientific disciplines.} We investigate if the changes we observe are different wrt. the scientific discipline (biology, computer science, medicine, psychology) that the finding originates from.
We report the percentage of finding pairs affected by critical distortions across the disciplines in Table~\ref{tab:change-types-comm-outlet}. Overall, we observe a similar level of  distortions across the disciplines biology, medicine and psychology. Findings from computer science show slightly different distortions: while they show increases in causal claim strength and increases in certainty more frequently compared to the other disciplines, the findings are generalized less often and they are less affected by increases in sensationalism compared to other disciplines.    

\begin{figure}
    \centering
    \includegraphics[scale=.4]{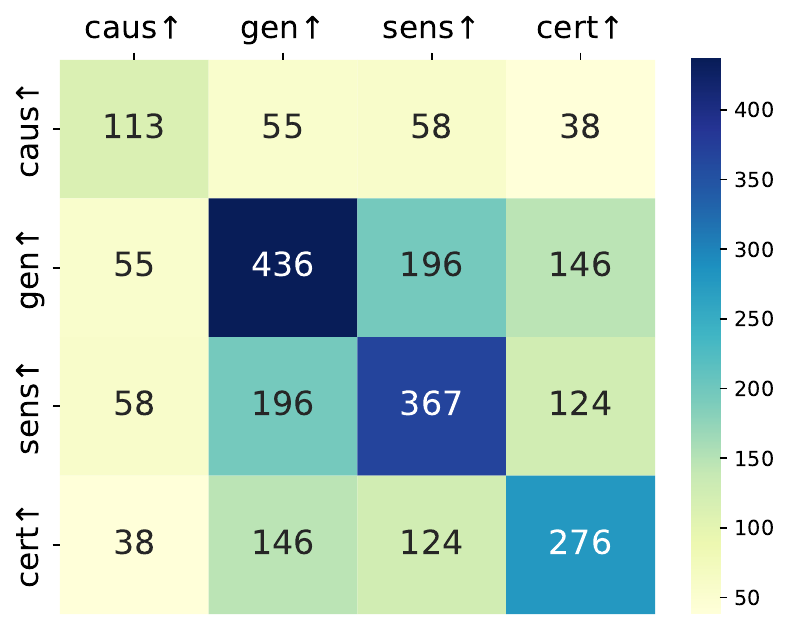}
    \caption{Co-occurrence matrix of critical distortions. Diagonals represent the number of paired findings affected by a particular distortion. All other counts represent distortion co-occurrences.}
    \label{fig:change-cooccurence}
\end{figure}

\paragraph{Co-occurrence of changes.}
We analyze the co-occurrence of critical distortion labels to understand potential connections between them. 
For every paired finding, each critical change is a binary variable that is True when the pair is affected by the change, and False if not. We plot the co-occurrence of these variables in Fig.~\ref{fig:change-cooccurence}.  
The distortions which co-occur most frequently are generalizations and increased sensationalism (196 instances). This is intuitive as findings may be sensationalist, because they convey broad or generalized claims. Similarly, increased certainty frequently co-occurs with generalization (146 instances) as well as increases sensationalism (124). Findings that convey strong claims with heightened certainty may be perceived as sensational and vice versa.

\section{Experiments}
\label{experiments}
In RQ2, we investigate how reliably we can detect information changes automatically. For all modeling experiments, we collapse the causality labels \textit{Expl. states: no relation} and \textit{No relation mentioned} into the \textit{Unclear relation} instances, and the certainty labels \textit{Somewhat uncertain} into the \textit{Uncertain} instances. We experiment with two modeling approaches which we describe in the following. 

\subsection{Setup}
\subsubsection{Task-specific Models}
To establish baselines for automatically predicting fine-grained distortions of scientific findings, we fine-tune task-specific models for each distortion type. We model \textit{causality}, \textit{certainty} and \textit{generalization} as classification tasks and predicting sensationalism scores as a regression. All models obtain as input a finding and learn to predict the distortion label. For \textit{generalization} the input is a paired finding. For each task, we train a classifier/regressor on top of a transformer base model.
\paragraph{Experimental setting.}
We experiment with two base models (RoBERTa-base\footnote{\url{https://huggingface.co/roberta-base}} and SciBERT\footnote{\url{https://huggingface.co/allenai/scibert_scivocab_uncased}}) to understand if domain-specific pretraining data is beneficial for our tasks.
For details on model training refer to Appendix~\ref{appendix:experimental-setting}.
We train all models using a 80/20 train-test split of the dataset.

 \begin{table*}
        \centering \small
        \begin{tabular}{lrrrl|lrrl|rrrl|l}
            \toprule
            model
            & \multicolumn{4}{c}{causality}& 
            \multicolumn{4}{c}{certainty}&
            \multicolumn{4}{c}{generalization}&
            \multicolumn{1}{c}{sensational.}\\

        \cmidrule(lr){1-1}\cmidrule(lr){2-5} \cmidrule(lr){6-9} \cmidrule(lr){10-13}  \cmidrule(r){14-14}  
            
            %causal
            &cau& corr & uncl & m\F
            % certainty
            & c &s\_c &unc& m\F
            % general
            &same &r\_f  &p\_f& m\F 
            % sensational
             & $r$ \\  

             \cmidrule(lr){2-2}\cmidrule(lr){3-3} \cmidrule(lr){4-4} \cmidrule(lr){5-5}
\cmidrule(lr){6-6} \cmidrule(lr){7-7} \cmidrule(lr){8-8} \cmidrule(lr){9-9} \cmidrule(lr){10-10} \cmidrule(lr){11-11} \cmidrule(lr){12-12} \cmidrule(lr){13-13} \cmidrule(lr){14-14} 
            
            LLaMa-2 &
            % causal
            .52 & .42 & .43 & .46&
            % certainty
            .04 & .49 & .37 & .3
            % general
            &.37 &.04  & .26& .22
            % sensational
            & .24
            \\

            Mistral7B &
            % causal
            .47& .43& .25 &.38
             
            % certainty
            
             & .5 & .05 &.34 &.3
            % general
             & .04 & .62 &.32 &.33
            % sensational
            & .10
            \\

            MiXtral8x7B &
            % causal
            .49 &.38  &.23 &.4
            % certainty
            
             & .42 &.02  &.23 &.22
            % general
             & .32 & .47 & .10&.3
            % sensational
            & .57
            \\
\cmidrule(lr){2-14}
            RoBERTa &
            % causal
             .56&.62& .59& .57$_{\pm.01}$&
            % certainty
            .67& .48& .51& .59$_{\pm.04}$ &
            % general
             .32& .69 & .42 & .40$_{\pm.06}$
            % sensational
             &.61$_{\pm.02}$
            \\
            
            SciBERT &
            % causal
             .58& .57&.60&.54$_{\pm.04}$& 
            % certainty
            .70&.50& .50 & .53$_{\pm.02}$
            % general
             &.32& .72 & .49 & .47$_{\pm.04}$
            % sensational
             &.57$_{\pm.03}$
             \\

            \bottomrule
        \end{tabular}
        \caption{Performance for predicting distortion labels of a finding. We report per class \F scores, macro \F and correlation coefficients ($r$, where applicable). For few-shot prompting, we use LlaMa2-chat-hf-13B, Mistral7B and Mixtral8x7B. For fine-tuning, we use RoBERTa-base and SciBERT. For fine-tuning experiments, we report per class results of the best performing model; m\F scores denote avg. across 5 runs, including standard deviation.}
        \label{tab:results-label-prediction}
    \end{table*}

\paragraph{Evaluation.}
We evaluate the model performance using the task-specific evaluation metrics, i.e., macro \F and Pearson's $r$ (see Sec.~\ref{eval-metrics}).

\subsubsection{Few-shot Prompting}
With the recent paradigm shift to in-context learning using instruction-tuned large language models (LLMs), we investigate the extent to which an LLM predicts, i.e., generates the correct change type label when prompted with the same instructions as human annotators.
%.
\paragraph{Evaluation.}
To calculate performance metrics, we need to extract the label from the LLM's output. To this end, we assume the first mention of any label from the task-specific label space that is closest to the answer cue in the question to be the prediction. We evaluate\footnote{For \textit{sensationalism}, we report correlations for the full dataset as opposed to only for the test portion. We mimic the best-worst-scaling setup from annotation and obtain the sensationalism score from the identical set of quad-tuples to allow for direct result comparison.} the model performance using the task-specific evaluation metrics (see Sec.~\ref{eval-metrics}).
\paragraph{Experimental setting.}
We experiment with three open, instruction-tuned LLMs for the few-shot prompting, varying in model size and architecture: LLaMa-2~13B\footnote{\url{https://huggingface.co/meta-llama/Llama-2-13b-chat-hf}}~\citep{touvron2023llama}, Mistral~7B\footnote{\url{https://huggingface.co/mistralai/Mistral-7B-Instruct-v0.1}}~\citep{mistral7b} and Mixtral~8x7B\footnote{\url{https://huggingface.co/mistralai/Mixtral-8x7B-Instruct-v0.1}}~\citep{mixtral}. As prompts, we use the same task description and examples that we used to instruct the human annotators. We provide our prompt template in Fig.~\ref{fig:prompt-template}.

\subsection{Results: How reliably can we detect distortions automatically? (RQ2)}
Table~\ref{tab:results-label-prediction} shows the results. Across all tasks, we observe that fine-tuned models exceed the performance of the few-shot prompting setup. For \textit{causality}, the fine-tuned SciBERT achieves a macro \F of 0.54$_{\pm.04}$ (avg. across 5 seeds, subscript denotes standard dev.), while the best results from prompting (LLaMa) achieves an \F of 0.46. Similarly, for  \textit{certainty}, SciBERT achieves an avg. \F of 0.53$_{\pm.02}$, while all LLMs struggle to make meaningful predictions. 
For \textit{generalization}, the avg. performance of the fine-tuned SciBERT is at 0.47$_{\pm.04}$~\F. For \textit{sensationalism}, fine-tuning RoBERTa obtains the best results ($r=$.61$_{\pm.02}$). Notably, it is the only task in which the few-shot approach obtains comparable results. \textit{Sensationalism} is also the only task for which the general domain model (RoBERTa) outperforms SciBERT. We presume that this is because detecting sensationalism is not strictly tied to scientific language, while the other tasks benefit from more specialized knowledge.

Overall, our results show that predicting fine-grained information changes is a very challenging task. Task-specific models produce more reliable results, while few-shot prompting performs poorly, even with large SOTA models. LLMs do not appear to be able to leverage the same instructions as humans, indicating that additional prompt engineering or fine-tuning may be required to obtain stronger results.

\section{Understanding Distorted Science Communication  at Large}
\label{large-data-anlaysis}
To gauge critical information changes broadly, we use a large scale set of paper findings, news findings, and tweets and analyze the prevalence of distortions in science communication at large.
\paragraph{Data.} The initial dataset is collected by pairing scientific papers from the S2ORC dataset~\cite{DBLP:conf/acl/LoWNKW20} with news articles and tweets using Altmetric\footnote{\url{https://www.altmetric.com/}}, an aggregator of mentions of scientific papers online. We automatically identify the result descriptions and select for each news and tweet finding the paper finding with the highest information matching score~\citep{wright-et-al-2022} above 4. Sec.~\ref{appendix:filtering-large-scale} describes the filtering process in detail. This gives us a set of 35,150 findings paired between papers and news and 72,032 findings paired between papers and tweets.

\begin{figure}
    \centering
    \includegraphics[scale=.4]{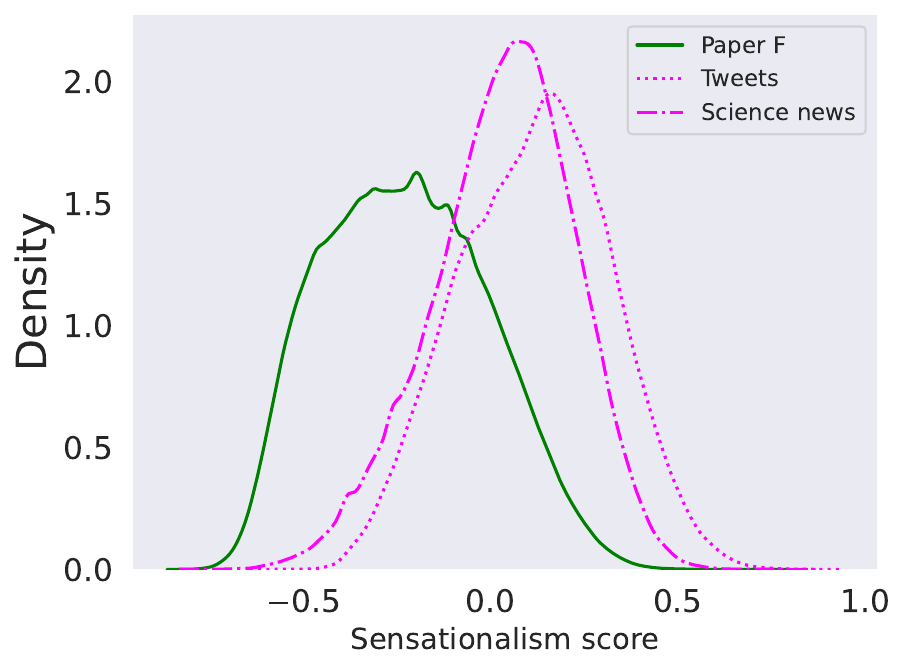}
    \caption{Density plot visualizing distribution of sensationalism scores across 1,655,570 paper, 422,626 news, and 356,275 tweet findings. Differences in the degree of sensationalism across different findings sources are statistically significant (see Fig. \ref{fig:sensationalism-significance} in Appendix \ref{sec:additional_figs}).}
    \label{fig:large-scale-sensational_density}
\end{figure}

Using the best performing models from Sec.\ref{experiments}, we estimate critical distortions wrt. \textit{causality},  \textit{certainty} and \textit{sensationalism}.\footnote{We exclude \textit{generalization} from this analysis because of the varied classification performance across target classes.} Our goal is to understand which type of reports --news vs. tweets-- are more susceptible to mis-reporting. 
\paragraph{Results.} Overall, we find that science communication on Twitter is more frequently affected by mis-reporting. Tweets show pronounced critical changes in causality (Fig.~\ref{fig:large-date_causal-tweets}, Appendix \ref{sec:additional_figs}), while the vast majority of findings reported in science news  accurately report the causal relation from the original finding (Fig.~\ref{fig:large-date_causal-news}, Appendix \ref{sec:additional_figs}). 
Science communicators on Twitter frequently overstate findings' certainty. Fig.~\ref{fig:large-data_certainty} in Appendix \ref{sec:additional_figs} shows changes in certainty levels. Notably, the vast majority of findings reported in tweets exhibit an increased level of certainty. This effect is less pronounced in the findings reported in science news.
Most notably, reported scientific findings are presented with heightened levels in sensationalism, both in tweets and science news. The density pot visualizing sensationalism scores in Fig.~\ref{fig:large-scale-sensational_density} shows a large shift of towards increased sensationalism for reported findings compared to their counterparts in the original papers.

\paragraph{Analysis.}
To validate the robustness of these results, we annotate 108 instances of the unlabeled data for the properties of \textit{causality} and \textit{certainty}. Sec.~\ref{appendix:validation-annotation} provides details on the annotation. Evaluating the predicted distortions against the human labels, we find that the results in this setting are robust (macro\F of .65 and .59 causal distortions and certainty) and on par with the classifiers’ performance on the original test set. Notably, given the relatively high precision for predicting the ``certain'' class (0.75) and the volume of reported findings which are predicted as such, these results provide further evidence that reports frequently overstate findings' certainty (Fig.~\ref{fig:large-data_certainty}). 
For \textit{sensationalism} we validate the results by analyzing if the difference in distributions that we observe in Fig.~\ref{fig:large-scale-sensational_density} is in fact significant. We see a large statistically significant effect~(Fig.~\ref{fig:sensationalism-significance}), indicating that different sources are potentially perceived as more or less sensational.

\section{Conclusion}
Given both the societal impact and growing volume of scientific information online, it is crucial to understand how this information is presented to the public. In this work, we lay a foundation for performing large scale analysis and automatic detection of several types of distortions in the reporting of scientific findings. 
We contribute the first dataset of multiple fine-grained distortions in science communication, allowing us to study how scientific findings are changed when reported to lay audiences. We show both that findings frequently undergo subtle distortions when reported, and that detecting these distortions automatically poses a challenging problem. We find that fine-tuning custom models consistently outperforms LLM prompting, presumably because the models are not able to effectively leverage the annotation instructions and examples we provide as prompts. Using our best baseline models, we study the prevalence of distortions in science communication at large, observing that scientific findings are potentially frequently subject to distortions in terms of causality, level of certainty, and how sensationally they are presented.

\section*{Limitations}
While we extend existing work wrt. the dimensions of distortions we study, this set may even be further extended. We focus on types of distortions which we consider to be applicable for a large set of disciplines, but there may be discipline-specific properties of misreporting that are out of scope of our current set of labels. Further, we focus on reports and scientific findings in English. Future work should extend this to other languages to understand the impact of the source language the  change types and their verbalization. 

Our annotation study shows that the properties we are investigating are highly complex and to a certain extent subjective as indicated by the mixed agreement scores. We argue that the concepts themselves are well defined, however, we hypothesize that the mixed inter-coder agreement is a result of the fact that concepts such as sensationalism and generalization are challenging to identify on the textual level, especially for the scientific domain. We account for this for example by choosing a best-worst-scaling setup for labeling sensationalism, however, this does not (and should not) fully remove the subjective nature of the task. Sensationalism for example can be encoded in a variety of linguistic cues, but it is also connected to certain topics: someone might perceive a finding about Covid vaccines to be more sensationalist as opposed to a different treatment, because it is more frequently the topic of recent discussion. This may lead to variance in the labels without one of them necessarily being incorrect. Further analysis and modeling of the factors that constitute the perception of each of these properties should be the focus of future work. 

With respect to our few-shot prompting experiments, we do not explore elaborate prompt engineering as the current work constitutes a baseline to explore in-context learning for our task. In the future, we aim to experiment with different prompts, and models and methods to extract labels from the generated output of the model as this is potentially error-prone.
Further, we currently model each distortion type separately, however, to a certain degree the properties we investigate are related to each other which a joint model may be able to leverage. We see this as an opportunity for future work.

For the large-scale analysis in Sec.~\ref{large-data-anlaysis}, the results have to be contextualized  with the performance of the models we use to automatically label the dataset. Our analyses validate that those findings are robust and serve as a starting point for large-scale analyses, showing that distortions are prevalent in science communication at large. However, future work is needed to determine their full extent with further developed models.

\section*{Ethical Considerations}
Inaccurate science reporting is a form of mis-information, our work can therefore contribute to detecting and counter-acting false information online.

This being said, while creating the resource to better understand this very task, annotators may be exposed to false information. We educate annotators about this possibility before they start the task. They can stop working on the task at any time. 

\section*{Acknowledgments}
This research has been conducted as part of the FIBISS project which is funded by the German Research Council (DFG, project number: KL~2869/5-1). 

% Entries for the entire Anthology, followed by custom entries
\bibliography{anthology,custom}

\begin{thebibliography}{39}
\expandafter\ifx\csname natexlab\endcsname\relax\def\natexlab#1{#1}\fi

\bibitem[{August et~al.(2020)August, Kim, Reinecke, and Smith}]{august-et-al-2020}
Tal August, Lauren Kim, Katharina Reinecke, and Noah~A. Smith. 2020.
\newblock \href {https://doi.org/10.18653/v1/2020.emnlp-main.429} {Writing strategies for science communication: Data and computational analysis}.
\newblock In \emph{Proceedings of the 2020 Conference on Empirical Methods in Natural Language Processing (EMNLP)}, pages 5327--5344, Online. Association for Computational Linguistics.

\bibitem[{Bagdon(2023)}]{bagdon-2023}
Christopher Bagdon. 2023.
\newblock You are and expert annotator: {A}utomated annotations with generative models and {BWS}.

\bibitem[{Beltagy et~al.(2019)Beltagy, Lo, and Cohan}]{DBLP:conf/emnlp/BeltagyLC19}
Iz~Beltagy, Kyle Lo, and Arman Cohan. 2019.
\newblock \href {https://doi.org/10.18653/V1/D19-1371} {Scibert: {A} pretrained language model for scientific text}.
\newblock In \emph{Proceedings of the 2019 Conference on Empirical Methods in Natural Language Processing and the 9th International Joint Conference on Natural Language Processing, {EMNLP-IJCNLP} 2019, Hong Kong, China, November 3-7, 2019}, pages 3613--3618. Association for Computational Linguistics.

\bibitem[{Boland et~al.(2022)Boland, Fafalios, Tchechmedjiev, Dietze, and Todorov}]{boland2022}
Katarina Boland, Pavlos Fafalios, Andon Tchechmedjiev, Stefan Dietze, and Konstantin Todorov. 2022.
\newblock Beyond facts--a survey and conceptualisation of claims in online discourse analysis.
\newblock \emph{Semantic Web -- Interoperability, Usability, Applicability}, 13(5):793--827.

\bibitem[{Bratton et~al.(2019)Bratton, Adams, Challenger, Boivin, Bott, Chambers, and Sumner}]{bratton2019association}
Luke Bratton, Rachel~C Adams, Aim{\'e}e Challenger, Jacky Boivin, Lewis Bott, Christopher~D Chambers, and Petroc Sumner. 2019.
\newblock The association between exaggeration in health-related science news and academic press releases: a replication study.
\newblock \emph{Wellcome open research}, 4.

\bibitem[{Canese and Weis(2013)}]{canese2013pubmed}
Kathi Canese and Sarah Weis. 2013.
\newblock {PubMed: the Bibliographic Database}.
\newblock \emph{The NCBI handbook}, 2(1).

\bibitem[{Dempster et~al.(2022)Dempster, Sutherland, and Keogh}]{dempster2022scientific}
Georgia Dempster, Georgina Sutherland, and Louise Keogh. 2022.
\newblock {Scientific Research in News Media: A Case Study of Misrepresentation, Sensationalism and Harmful Recommendations}.
\newblock \emph{Journal of Science Communication}, 21(1):A06.

\bibitem[{Fischhoff(2012)}]{fischhoff2012communicating}
Baruch Fischhoff. 2012.
\newblock Communicating uncertainty fulfilling the duty to inform.
\newblock \emph{Issues in Science and Technology}, 28(4):63--70.

\bibitem[{Guo et~al.(2022)Guo, Schlichtkrull, and Vlachos}]{guo-etal-2022-survey}
Zhijiang Guo, Michael Schlichtkrull, and Andreas Vlachos. 2022.
\newblock \href {https://doi.org/10.1162/tacl_a_00454} {A survey on automated fact-checking}.
\newblock \emph{Transactions of the Association for Computational Linguistics}, 10:178--206.

\bibitem[{Hart and Feldman(2016)}]{hart2016impact}
P~Sol Hart and Lauren Feldman. 2016.
\newblock The impact of climate change--related imagery and text on public opinion and behavior change.
\newblock \emph{Science Communication}, 38(4):415--441.

\bibitem[{Hovy et~al.(2013)Hovy, Berg-Kirkpatrick, Vaswani, and Hovy}]{hovy-etal-2013-learning}
Dirk Hovy, Taylor Berg-Kirkpatrick, Ashish Vaswani, and Eduard Hovy. 2013.
\newblock \href {https://aclanthology.org/N13-1132} {Learning whom to trust with {MACE}}.
\newblock In \emph{Proceedings of the 2013 Conference of the North {A}merican Chapter of the Association for Computational Linguistics: Human Language Technologies}, pages 1120--1130, Atlanta, Georgia. Association for Computational Linguistics.

\bibitem[{Hripcsak and Rothschild(2005)}]{hripcsak_agreement_2005}
George Hripcsak and Adam~S. Rothschild. 2005.
\newblock \href {https://doi.org/10.1197/jamia.M1733} {Agreement, the f-measure, and reliability in information retrieval}.
\newblock \emph{Journal of the American Medical Informatics Association}, 12(3):296--298.
\newblock \_eprint: https://academic.oup.com/jamia/article-pdf/12/3/296/2429751/12-3-296.pdf.

\bibitem[{Jiang et~al.(2023)Jiang, Sablayrolles, Mensch, Bamford, Chaplot, de~las Casas, Bressand, Lengyel, Lample, Saulnier, Lavaud, Lachaux, Stock, Scao, Lavril, Wang, Lacroix, and Sayed}]{mistral7b}
Albert~Q. Jiang, Alexandre Sablayrolles, Arthur Mensch, Chris Bamford, Devendra~Singh Chaplot, Diego de~las Casas, Florian Bressand, Gianna Lengyel, Guillaume Lample, Lucile Saulnier, Lélio~Renard Lavaud, Marie-Anne Lachaux, Pierre Stock, Teven~Le Scao, Thibaut Lavril, Thomas Wang, Timothée Lacroix, and William~El Sayed. 2023.
\newblock \href {http://arxiv.org/abs/2310.06825} {Mistral 7b}.

\bibitem[{Jiang et~al.(2024)Jiang, Sablayrolles, Roux, Mensch, Savary, Bamford, Chaplot, de~las Casas, Hanna, Bressand, Lengyel, Bour, Lample, Lavaud, Saulnier, Lachaux, Stock, Subramanian, Yang, Antoniak, Scao, Gervet, Lavril, Wang, Lacroix, and Sayed}]{mixtral}
Albert~Q. Jiang, Alexandre Sablayrolles, Antoine Roux, Arthur Mensch, Blanche Savary, Chris Bamford, Devendra~Singh Chaplot, Diego de~las Casas, Emma~Bou Hanna, Florian Bressand, Gianna Lengyel, Guillaume Bour, Guillaume Lample, Lélio~Renard Lavaud, Lucile Saulnier, Marie-Anne Lachaux, Pierre Stock, Sandeep Subramanian, Sophia Yang, Szymon Antoniak, Teven~Le Scao, Théophile Gervet, Thibaut Lavril, Thomas Wang, Timothée Lacroix, and William~El Sayed. 2024.
\newblock \href {http://arxiv.org/abs/2401.04088} {Mixtral of experts}.

\bibitem[{Kiritchenko and Mohammad(2017)}]{kiritchenko-mohammad-2017-best}
Svetlana Kiritchenko and Saif Mohammad. 2017.
\newblock \href {https://doi.org/10.18653/v1/P17-2074} {Best-worst scaling more reliable than rating scales: A case study on sentiment intensity annotation}.
\newblock In \emph{Proceedings of the 55th Annual Meeting of the Association for Computational Linguistics (Volume 2: Short Papers)}, pages 465--470, Vancouver, Canada. Association for Computational Linguistics.

\bibitem[{Kuehne and Olden(2015)}]{kuehne-olden-2015}
Lauren~M Kuehne and Julian~D Olden. 2015.
\newblock Lay summaries needed to enhance science communication.
\newblock \emph{Proceedings of the National Academy of Sciences}, 112(12):3585--3586.

\bibitem[{Kuru et~al.(2021)Kuru, Stecula, Lu, Ophir, Chan, Winneg, Hall~Jamieson, and Albarrac{\'\i}n}]{kuru2021effects}
Ozan Kuru, Dominik Stecula, Hang Lu, Yotam Ophir, Man-pui~Sally Chan, Ken Winneg, Kathleen Hall~Jamieson, and Dolores Albarrac{\'\i}n. 2021.
\newblock The effects of scientific messages and narratives about vaccination.
\newblock \emph{PLoS One}, 16(3):e0248328.

\bibitem[{Kuznetsov et~al.(2022)Kuznetsov, Buchmann, Eichler, and Gurevych}]{DBLP:journals/coling/KuznetsovBEG22}
Ilia Kuznetsov, Jan Buchmann, Max Eichler, and Iryna Gurevych. 2022.
\newblock \href {https://doi.org/10.1162/COLI\_A\_00455} {{Revise and Resubmit: An Intertextual Model of Text-based Collaboration in Peer Review}}.
\newblock \emph{Comput. Linguistics}, 48(4):949--986.

\bibitem[{Lawrence and Reed(2019)}]{lawrence-reed-2019-argument}
John Lawrence and Chris Reed. 2019.
\newblock \href {https://doi.org/10.1162/coli_a_00364} {Argument mining: A survey}.
\newblock \emph{Computational Linguistics}, 45(4):765--818.

\bibitem[{Liu et~al.(2019)Liu, Ott, Goyal, Du, Joshi, Chen, Levy, Lewis, Zettlemoyer, and Stoyanov}]{DBLP:journals/corr/abs-1907-11692}
Yinhan Liu, Myle Ott, Naman Goyal, Jingfei Du, Mandar Joshi, Danqi Chen, Omer Levy, Mike Lewis, Luke Zettlemoyer, and Veselin Stoyanov. 2019.
\newblock \href {http://arxiv.org/abs/1907.11692} {Roberta: {A} robustly optimized {BERT} pretraining approach}.
\newblock \emph{CoRR}, abs/1907.11692.

\bibitem[{Lo et~al.(2020)Lo, Wang, Neumann, Kinney, and Weld}]{DBLP:conf/acl/LoWNKW20}
Kyle Lo, Lucy~Lu Wang, Mark Neumann, Rodney Kinney, and Daniel~S. Weld. 2020.
\newblock \href {https://doi.org/10.18653/V1/2020.ACL-MAIN.447} {{S2ORC:} the semantic scholar open research corpus}.
\newblock In \emph{Proceedings of the 58th Annual Meeting of the Association for Computational Linguistics, {ACL} 2020, Online, July 5-10, 2020}, pages 4969--4983. Association for Computational Linguistics.

\bibitem[{McHugh(2012)}]{McHugh2012}
Mary~L McHugh. 2012.
\newblock Interrater reliability: the kappa statistic.
\newblock \emph{Biochem Med (Zagreb)}, 22(3):276--282.

\bibitem[{Mohr et~al.(2022)Mohr, W{\"{u}}hrl, and Klinger}]{DBLP:conf/lrec/MohrWK22}
Isabelle Mohr, Amelie W{\"{u}}hrl, and Roman Klinger. 2022.
\newblock \href {https://aclanthology.org/2022.lrec-1.26} {Covert: {A} corpus of fact-checked biomedical {COVID-19} tweets}.
\newblock In \emph{Proceedings of the Thirteenth Language Resources and Evaluation Conference, {LREC} 2022, Marseille, France, 20-25 June 2022}, pages 244--257. European Language Resources Association.

\bibitem[{Pei et~al.(2022)Pei, Ananthasubramaniam, Wang, Zhou, Dedeloudis, Sargent, and Jurgens}]{pei2022}
Jiaxin Pei, Aparna Ananthasubramaniam, Xingyao Wang, Naitian Zhou, Apostolos Dedeloudis, Jackson Sargent, and David Jurgens. 2022.
\newblock Potato: The portable text annotation tool.
\newblock In \emph{Proceedings of the 2022 Conference on Empirical Methods in Natural Language Processing: System Demonstrations}.

\bibitem[{Pei and Jurgens(2021{\natexlab{a}})}]{pei-jurgens-2021}
Jiaxin Pei and David Jurgens. 2021{\natexlab{a}}.
\newblock \href {https://doi.org/10.18653/v1/2021.emnlp-main.784} {Measuring sentence-level and aspect-level (un)certainty in science communications}.
\newblock In \emph{Proceedings of the 2021 Conference on Empirical Methods in Natural Language Processing}, pages 9959--10011, Online and Punta Cana, Dominican Republic. Association for Computational Linguistics.

\bibitem[{Pei and Jurgens(2021{\natexlab{b}})}]{pei-jurgens-2021-measuring}
Jiaxin Pei and David Jurgens. 2021{\natexlab{b}}.
\newblock \href {https://doi.org/10.18653/v1/2021.emnlp-main.784} {Measuring sentence-level and aspect-level (un)certainty in science communications}.
\newblock In \emph{Proceedings of the 2021 Conference on Empirical Methods in Natural Language Processing}, pages 9959--10011, Online and Punta Cana, Dominican Republic. Association for Computational Linguistics.

\bibitem[{Ransohoff and Ransohoff(2001)}]{ransohoff2001sensationalism}
David~F Ransohoff and Richard~M Ransohoff. 2001.
\newblock {Sensationalism in the Media: When Scientists and Journalists May be Complicit Collaborators.}
\newblock \emph{Effective clinical practice}, 4(4).

\bibitem[{Salita(2015)}]{salita2015writing}
Joselita~T Salita. 2015.
\newblock Writing for lay audiences: A challenge for scientists.
\newblock \emph{Medical Writing}, 24(4):183--189.

\bibitem[{Sumner et~al.(2014)Sumner, Vivian-Griffiths, Boivin, Williams, Venetis, Davies, Ogden, Whelan, Hughes, Dalton et~al.}]{sumner2014association}
Petroc Sumner, Solveiga Vivian-Griffiths, Jacky Boivin, Andy Williams, Christos~A Venetis, Aim{\'e}e Davies, Jack Ogden, Leanne Whelan, Bethan Hughes, Bethan Dalton, et~al. 2014.
\newblock The association between exaggeration in health related science news and academic press releases: retrospective observational study.
\newblock \emph{Bmj}, 349.

\bibitem[{Tichenor et~al.(1970)Tichenor, Olien, Harrison, and Donohue}]{tichenor1970mass}
Phillip~J Tichenor, Clarice~N Olien, Annette Harrison, and George Donohue. 1970.
\newblock {Mass Communication Systems and Communication Accuracy in Science News Reporting}.
\newblock \emph{Journalism Quarterly}, 47(4):673--683.

\bibitem[{Touvron et~al.(2023)Touvron, Martin, Stone, Albert, Almahairi, Babaei, Bashlykov, Batra, Bhargava, Bhosale, Bikel, Blecher, Ferrer, Chen, Cucurull, Esiobu, Fernandes, Fu, Fu, Fuller, Gao, Goswami, Goyal, Hartshorn, Hosseini, Hou, Inan, Kardas, Kerkez, Khabsa, Kloumann, Korenev, Koura, Lachaux, Lavril, Lee, Liskovich, Lu, Mao, Martinet, Mihaylov, Mishra, Molybog, Nie, Poulton, Reizenstein, Rungta, Saladi, Schelten, Silva, Smith, Subramanian, Tan, Tang, Taylor, Williams, Kuan, Xu, Yan, Zarov, Zhang, Fan, Kambadur, Narang, Rodriguez, Stojnic, Edunov, and Scialom}]{touvron2023llama}
Hugo Touvron, Louis Martin, Kevin Stone, Peter Albert, Amjad Almahairi, Yasmine Babaei, Nikolay Bashlykov, Soumya Batra, Prajjwal Bhargava, Shruti Bhosale, Dan Bikel, Lukas Blecher, Cristian~Canton Ferrer, Moya Chen, Guillem Cucurull, David Esiobu, Jude Fernandes, Jeremy Fu, Wenyin Fu, Brian Fuller, Cynthia Gao, Vedanuj Goswami, Naman Goyal, Anthony Hartshorn, Saghar Hosseini, Rui Hou, Hakan Inan, Marcin Kardas, Viktor Kerkez, Madian Khabsa, Isabel Kloumann, Artem Korenev, Punit~Singh Koura, Marie-Anne Lachaux, Thibaut Lavril, Jenya Lee, Diana Liskovich, Yinghai Lu, Yuning Mao, Xavier Martinet, Todor Mihaylov, Pushkar Mishra, Igor Molybog, Yixin Nie, Andrew Poulton, Jeremy Reizenstein, Rashi Rungta, Kalyan Saladi, Alan Schelten, Ruan Silva, Eric~Michael Smith, Ranjan Subramanian, Xiaoqing~Ellen Tan, Binh Tang, Ross Taylor, Adina Williams, Jian~Xiang Kuan, Puxin Xu, Zheng Yan, Iliyan Zarov, Yuchen Zhang, Angela Fan, Melanie Kambadur, Sharan Narang, Aurelien Rodriguez, Robert Stojnic, Sergey Edunov, and Thomas
  Scialom. 2023.
\newblock \href {http://arxiv.org/abs/2307.09288} {Llama 2: Open foundation and fine-tuned chat models}.

\bibitem[{Trienes et~al.(2024)Trienes, Joseph, Schlötterer, Seifert, Lo, Xu, Wallace, and Li}]{trienes2024}
Jan Trienes, Sebastian Joseph, Jörg Schlötterer, Christin Seifert, Kyle Lo, Wei Xu, Byron~C. Wallace, and Junyi~Jessy Li. 2024.
\newblock \href {http://arxiv.org/abs/2401.16475} {Infolossqa: Characterizing and recovering information loss in text simplification}.

\bibitem[{Vladika and Matthes(2023)}]{vladika-matthes-2023-scientific}
Juraj Vladika and Florian Matthes. 2023.
\newblock \href {https://doi.org/10.18653/v1/2023.findings-acl.387} {Scientific fact-checking: A survey of resources and approaches}.
\newblock In \emph{Findings of the Association for Computational Linguistics: ACL 2023}, pages 6215--6230, Toronto, Canada. Association for Computational Linguistics.

\bibitem[{Wadden et~al.(2020)Wadden, Lin, Lo, Wang, van Zuylen, Cohan, and Hajishirzi}]{DBLP:conf/emnlp/WaddenLLWZCH20}
David Wadden, Shanchuan Lin, Kyle Lo, Lucy~Lu Wang, Madeleine van Zuylen, Arman Cohan, and Hannaneh Hajishirzi. 2020.
\newblock \href {https://doi.org/10.18653/V1/2020.EMNLP-MAIN.609} {Fact or fiction: Verifying scientific claims}.
\newblock In \emph{Proceedings of the 2020 Conference on Empirical Methods in Natural Language Processing, {EMNLP} 2020, Online, November 16-20, 2020}, pages 7534--7550. Association for Computational Linguistics.

\bibitem[{Wright and Augenstein(2021{\natexlab{a}})}]{DBLP:conf/acl/WrightA21}
Dustin Wright and Isabelle Augenstein. 2021{\natexlab{a}}.
\newblock \href {https://doi.org/10.18653/V1/2021.FINDINGS-ACL.157} {{CiteWorth: Cite-Worthiness Detection for Improved Scientific Document Understanding}}.
\newblock In \emph{Findings of the Association for Computational Linguistics: {ACL/IJCNLP} 2021, Online Event, August 1-6, 2021}, volume {ACL/IJCNLP} 2021 of \emph{Findings of {ACL}}, pages 1796--1807. Association for Computational Linguistics.

\bibitem[{Wright and Augenstein(2021{\natexlab{b}})}]{wright-augenstein-2021}
Dustin Wright and Isabelle Augenstein. 2021{\natexlab{b}}.
\newblock \href {https://doi.org/10.18653/v1/2021.emnlp-main.845} {Semi-supervised exaggeration detection of health science press releases}.
\newblock In \emph{Proceedings of the 2021 Conference on Empirical Methods in Natural Language Processing}, pages 10824--10836, Online and Punta Cana, Dominican Republic. Association for Computational Linguistics.

\bibitem[{Wright et~al.(2022)Wright, Pei, Jurgens, and Augenstein}]{wright-et-al-2022}
Dustin Wright, Jiaxin Pei, David Jurgens, and Isabelle Augenstein. 2022.
\newblock \href {https://aclanthology.org/2022.emnlp-main.117} {Modeling information change in science communication with semantically matched paraphrases}.
\newblock In \emph{Proceedings of the 2022 Conference on Empirical Methods in Natural Language Processing}, pages 1783--1807, Abu Dhabi, United Arab Emirates. Association for Computational Linguistics.

\bibitem[{Yu et~al.(2019)Yu, Li, and Wang}]{yu-etal-2019}
Bei Yu, Yingya Li, and Jun Wang. 2019.
\newblock \href {https://doi.org/10.18653/v1/D19-1473} {Detecting causal language use in science findings}.
\newblock In \emph{Proceedings of the 2019 Conference on Empirical Methods in Natural Language Processing and the 9th International Joint Conference on Natural Language Processing (EMNLP-IJCNLP)}, pages 4664--4674, Hong Kong, China. Association for Computational Linguistics.

\bibitem[{Yu et~al.(2020)Yu, Wang, Guo, and Li}]{yu-etal-2020-measuring}
Bei Yu, Jun Wang, Lu~Guo, and Yingya Li. 2020.
\newblock \href {https://doi.org/10.18653/v1/2020.coling-main.427} {Measuring correlation-to-causation exaggeration in press releases}.
\newblock In \emph{Proceedings of the 28th International Conference on Computational Linguistics}, pages 4860--4872, Barcelona, Spain (Online). International Committee on Computational Linguistics.

\end{thebibliography}

\appendix

\section{Appendix}
\label{sec:appendix}

\subsection{Annotation}
\label{appendix:annotation}

\subsubsection{Setting}
\label{appendix:annotation-setting}
\paragraph{Participant filtering.} To qualify for our study, participants have to have an undergrad degree in one of the following subjects:
CS: Computer Science, Computing (IT), Mathematics, Science;
Biology \& Medicine: Biochemistry (Molecular and Cellular), Biological Sciences, Biology, Biomedical Sciences, Chemistry, Health and Medicine, Medicine;
Psychology: Psychology. Using Prolific's `Balanced sample' option, we rely on the platform to distribute the study evenly to male and female participants. Participants are required to be fluent in English. 
\paragraph{Payment.} All studies are designed to take approx. 13 minutes. Annotators are payed \textsterling 1.95 per study. This amounts to \textsterling 9 per hour which \textsc{Prolific} recommends as a fair compensation.
\paragraph{Number of annotations.} For the change types \emph{causality}, \emph{certainty}, and \emph{generalization}, we collect 3 sets of annotations for every instance. For \emph{sensationalism}, we generate 1.5$N$ quad-tuples, $N$ being the total number of findings to be labeled, and collect 2 sets of annotations for the resulting quad-tuples.

\subsubsection{Tasks}
\label{appendix:annotation-task}
Table \ref{table:annotation-setup} provides an overview of the annotation tasks.
\begin{table*}[t]
  \centering \small
  
  \begin{tabularx}{\linewidth}{lXXXX}
    \toprule
     & Causality &Certainty& Generalization & Sensationalism \\
    \cmidrule(lr){2-2} \cmidrule(lr){3-3} \cmidrule(lr){4-4} \cmidrule(lr){5-5}
    setup & classification & rating scale & comp. classification & best-worst-scaling \\
    
    task 
    &In the finding, what type of causal relationship is described?
    &How do you rate the level of certainty used to describe the finding?
    &Which finding is more general?
    &Which of the findings is the least/most sensational? \\
    
    label space 
    &Causation, Correlation, Expl. states: no relation, No relation mentioned 
    & Certain, Somewhat certain, Somewhat uncertain, Uncertain
    & Reported Finding, Paper Finding, Same level of generality
    & Most sensational, Least sensational \\

    aggregation & MACE & MACE  & MACE & count-based BWS-score\\
    
    \bottomrule

  \end{tabularx}
  \caption{Overview of annotation tasks, annotation setup, label and aggregation strategies. MACE: \citet{hovy-etal-2013-learning}, BWS-score~\citep{kiritchenko-mohammad-2017-best}: real-valued score obtained from the best-worst scaling.}
  \label{table:annotation-setup}
\end{table*}

\paragraph{Causality}
Annotators are provided the following task description and examples: ``Identify what type of causal relationship is described in a finding. Causality describes a cause-effect relationship between two things, variables, agents etc. A (directly) causes outcome B. Correlation describes relationships where two actions relate to each other, but one is not necessarily the effect or outcome of the other. Cues for causality are: \textit{cause, direct connection, result in, lead to, trigger, produce, increase, decrease}. Cues for correlation: \textit{associated with, association, connection, correlated with, linked to}. When in doubt or not clearly stated as a causal relation, it's usually a correlation.
Let's go over some examples:
FINDING 1: Low vitamin D levels cause tiredness.
FINDING 2: Exposure to traffic noise at the office increases stress levels.
→ Both examples describe a causal relationship: The cause A (low vitamin D, traffic noise) causes outcome B (tiredness, increased stress level).
Compared to that, this one describes a correlation:
FINDING 3: Low Vitamin D levels are associated with tiredness.
FINDING 4: Stress levels are higher in offices exposed to traffic noise.
→ Both examples describe a correlation. In both sentences the variables are related or associated to each other, but there it is unclear if one is the direct cause of the other.
Sometimes no relation is stated:
FINDING 5: We find evidence of biases across the majority of languages.
→ This finding presents a summary in which no causal relation or correlation is stated."

The labels are defined as follows; they follow~\citet{sumner2014association} and annotators see them as label descriptions in the annotation environment: 
\begin{itemize}
    \item No mention of a relation: No mention of a relation of any kind. E.g., if the finding is a summary such as \textit{We find evidence of biases across all languages}
\item Correlation: The paper finding describes a correlation between two elements. E.g., \textit{Vitamin D levels are associated with extensive tiredness}
\item Causation: The paper finding describes a causal relation between two elements. E.g., \textit{Low vitamin D levels cause extensive tiredness} or \textit{Tiredness might be caused by lack of vitamin D.}
\item Explicitly states: no relation: The finding explicitly states that there is no relation between the two elements. E.g., \textit{We find no evidence that vitamin D levels are associated with tiredness.}
\end{itemize}

\paragraph{Certainty}
Annotators are provided the following task description and examples: ``Rate the level of certainty that the author uses to describe a finding. Certainty means having complete confidence in something without any doubts. Uncertainty is the opposite: it refers to a state of doubt, lack of confidence, or absence of complete knowledge about something. Both can be expressed with respect to various aspects. Look out for: (un)certainty towards specific numbers/quantities: \textit{They found that approximately 50\% of participants\ldots}, the extent to which a given finding applies: \textit{The effect was mainly observed for teenagers.}), the probability that something applies, occurs or is associated: \textit{possibly associated with}, hedging words: \textit{seem, tend, appear to be, may, potentially, suggest, perhaps}. 
Let's go over some examples:
FINDING 1: Now there is clear evidence that sunscreen prevents skin damage.
→ The description of the finding is very certain. No hedges or other indicators of uncertainty.
Compared to that, this one is slightly less certain:
FINDING 2: New study shows that sunscreen can prevent skin damage.
→ The description of the finding is pretty certain. The use of \textit{can} indicates that the finding is limited in some way, but overall the finding is presented to be mostly certain.
Let's look at some examples that express uncertainty:
FINDING 3: New study suggests that sunscreen could prevent skin damage.
→ The finding is described to be pretty uncertain. The use of \textit{could} and \textit{suggests} are indicators that the findings are preliminary or very limited with regards to how impactful they may be.
Let's go all the way to an uncertain finding:
FINDING 4: Study presents potential indicators that sunscreen might have positive effects in preventing skin damage.
→ The finding is described to be very uncertain. It is stated that the results are indicators instead of a definite explanation. The use of the word \textit{might} emphasizes the uncertainty of the finding."

Annotators are provided the following label descriptions in the annotation environment:
\begin{itemize}
    \item Uncertain: E.g., 'Sunscreen might prevent skin cancer.' or 'Overconsumption of sugar may have negative effects on health.', 'Further research is necessary to understand\ldots'
    \item Somewhat uncertain: E.g., 'Sunscreen could prevent skin cancer.' or 'Overconsumption of sugar can cause diabetes.', 'The functionality possibly depends on\ldots'
    \item Somewhat certain: E.g., 'Sunscreen can prevent skin cancer.' or 'The analysis suggests that papers with short titles receive more citations'
    \item Certain: E.g., 'Sunscreen prevents skin cancer.' or 'Papers with shorter titles get more citations', '\ldots meaning that this treatment should be used\ldots'
\end{itemize}

\paragraph{Generalization}
Annotators are provided the following task description and examples: ``Identify if Finding A generalizes the results from Finding B. Generalizations claim something is always true, even if something is only valid in certain instances or occasionally.
Let's go over an example:
Read Finding A.
FINDING A: Parent conversations with children about their weight connected to disordered eating
Compare that to Finding B:
FINDING B: Disordered eating was more prevalent in children whose fathers engaged in weight conversations.
→ Finding B specifies that it is conversation with fathers which were investigated. Finding A generalizes the statement from fathers to all parents. In the study, you are tasked to decide which of the findings is more general: Here, the correct solution is Finding A.
Let's look at another example:
FINDING A: Magnesium potentially has many health benefits.
FINDING B: Increasing dietary magnesium intake is associated with a reduced risk of stroke and heart failure.
→ Here, Finding B specifies they researched dietary magnesium and the medical conditions that are affected. Finding A is more general, the correct solution is therefore Finding A.
Both findings can be on the same level of generality:
FINDING A: Dietary magnesium potentially has health benefits.
FINDING B: We show that dietary magnesium had a positive influence on the participants' overall health. The correct
solution is They are at the same level of generality."

Annotators are provided the following label descriptions in the annotation environment:
\begin{itemize}
    \item Finding A: E.g., Finding A discusses a general finding about diabetes, while the report discusses the finding for a specific demographic only. Or generalizing from a specified set of medical conditions (`reduced risk of stroke, heart failure, diabetes') to a general statement (`has health benefits').
    \item Finding B: E.g., Finding B generalizes a finding about diabetes type 2 in seniors to all people with diabetes. Or generalizing from a specified set of medical conditions (`reduced risk of stroke, heart failure, diabetes') to a general statement (`has health benefits').
    \item They are at the same level of generality: Finding B accurately reports the Finding A with regards to generality.
\end{itemize}

\paragraph{Sensationalism}
Annotators are provided the following task description and examples: ``Rate how sensational the language of a finding is. Sensational text intends to spark the interest of a reader, make them curious or elicit an emotional reaction. Cues for sensationalism could be: dramatic, urgent, exaggerated language: \textit{life-changing}, \textit{unparalleled performance}, \textit{revolutionary}, \textit{transformative}, \textit{miracle treatment} or use informal or colloquial language: \textit{amps up the efficiency}, \textit{They ran some solid experiments to back this up}.
Let's go over an example:
FINDING A: Looks like vitamin D intake influences stress levels: New study by @username.
FINDING B: Urban Green Spaces and Mental Health: A Positive Correlation Revealed.
FINDING C: Exciting new research suggests that upping your daily step count could be a simple solution to alleviate insomnia.
FINDING D: We observe improved plant growth through positive human energy in our controlled study setup.
Read all findings carefully. Some of the text contain sensational or entertaining language like \textit{revealed}, \textit{exciting}, or \textit{simple solution}. Based on that you determine which finding uses the most sensational language and which one the least: Here, Finding D is the LEAST sensational. FINDING C is the MOST sensational."

Annotators are provided the following label descriptions in the annotation environment:
\begin{itemize}
    \item A: Finding A is the least sensational.
    \item B: Finding B is the least sensational.
    \item C: Finding C is the least sensational.
    \item D: Finding D is the least sensational.
\end{itemize}
and 
\begin{itemize}
    \item A: Finding A is the most sensational.
    \item B: Finding B is the most sensational.
    \item C: Finding C is the most sensational.
    \item D: Finding D is the most sensational.
\end{itemize}

\subsection{Experiments}
\subsubsection{Experimental setting}
\label{appendix:experimental-setting}
\paragraph{Fine-tuning task-specific models.} We train all models using a Nvidia Titan-RTX GPU. We train for 5 epochs with a learning rate of 2e-5, a batch size of 8, 200 warmup steps, a weight decay of 0.01. We use the Adamw optimizer.
\paragraph{Few-shot prompting.} For each input prompt, we generate the output sequence by using top k sampling with k$=$10. We set the maximum number of generated tokens to 200. We use a NVIDIA Tesla V100 GPU to generate the responses with the LLaMa model. Generating the responses for the full dataset takes approx. 7 hours. For the Mistral7B model, generating all responses takes approx. 5 hours on an Nvidia GeForce RTX A6000 GPU. For Mixtral8x7B, generation takes approx. 20 hours split across 5 Nvidia GeForce RTX A6000 GPUs. We wrap each prompt with \textsc{[INST][/INST]} tags to mimic the chat format from pre-training and include a system prompt that instructs the model to `act' like a reliable annotator.
\subsubsection{Prompts}
\label{appendix:prompts}
Fig. \ref{fig:prompt-template} shows our prompt template.
For the BWS-prompts, we follow \citet{bagdon-2023} who experiment with automatically generating training data using best-worst-scaling.
We provide the task-specific prompts in the supplementary material.

\begin{figure}
\small
\begin{tabularx}{\linewidth}{X}
\hline
You are a reliable annotator in an annotation study. You studied \{\textsc{scientific discipline}\}\\
\{\textsc{task description}\}\\
\{\textsc{task examples}\}\\
Now consider the following finding:\\
\{\textsc{finding, finding pair, quad-tuple}\}
\{\textsc{question}\}\\
What is the correct solution? Choose one option. Do not repeat the findings.\\ 
\bottomrule
\end{tabularx}

    \caption{Prompt template. We provide the instantiated prompts along with the LLM-specific system prompts and markup in the supplementary material.}
    \label{fig:prompt-template}
\end{figure}

\subsection{Additional Analyses}
\label{sec:additional_figs}

\subsubsection{Filtering Process}
\label{appendix:filtering-large-scale}
The initial dataset is collected by pairing scientific papers from the S2ORC dataset~\cite{DBLP:conf/acl/LoWNKW20} with news articles and tweets using Altmetric, an aggregator of mentions of scientific papers online.\footnote{\url{https://www.altmetric.com/}} The scientific papers and news articles are initially parsed to predict which sentences correspond to either a result or conclusion using a RoBERTa model~\cite{DBLP:journals/corr/abs-1907-11692}\footnote{roberta-base} trained on 200K paper abstracts from PubMed that are self-labeled with paper section categories~\cite{canese2013pubmed}. We then pass all pairs of conclusion and result sentences from papers with conclusion and result sentences from news articles and tweets through the model from~\citet{wright-et-al-2022-modeling} to measure the similarity of the pair of findings, and select for each news and tweet finding the paper finding with the highest IMS above 4.

\subsubsection{Distortions in Large Scale Data}
\begin{figure}
    \begin{subfigure}{0.5\columnwidth}
        \centering
        \includegraphics[width=1\linewidth]{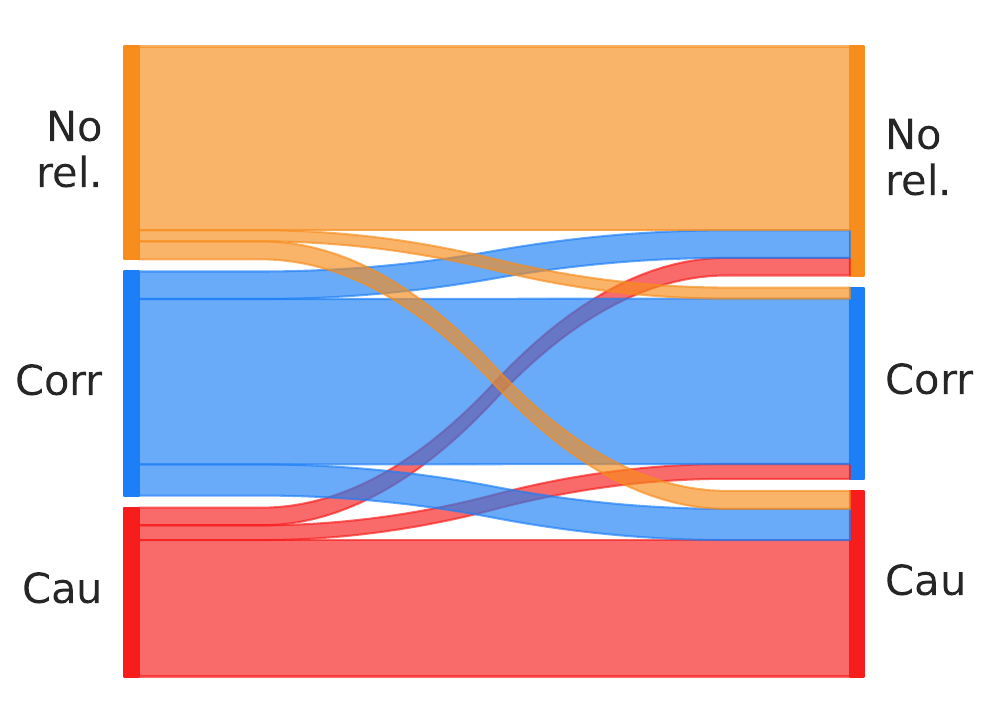}
        \caption{Science news}
        \label{fig:large-date_causal-news}
    \end{subfigure}%
    \begin{subfigure}{0.5\columnwidth}
        \centering
        \includegraphics[width=\linewidth]{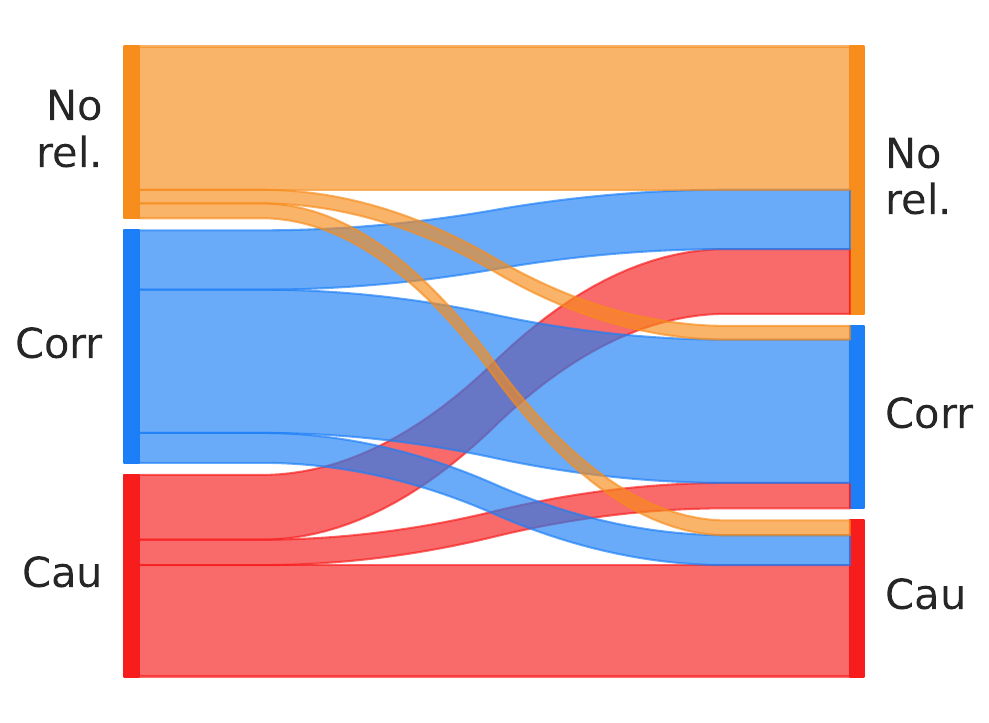}
        \caption{Tweets}
        \label{fig:large-date_causal-tweets}
    \end{subfigure}
    \caption{Changes in causality across paired findings from science news (35,150) and  tweets (72,032).}
    \label{fig:large-data_causal}
\end{figure}

\begin{figure}
    \begin{subfigure}{0.5\columnwidth}
        \centering
        \includegraphics[width=1\linewidth]{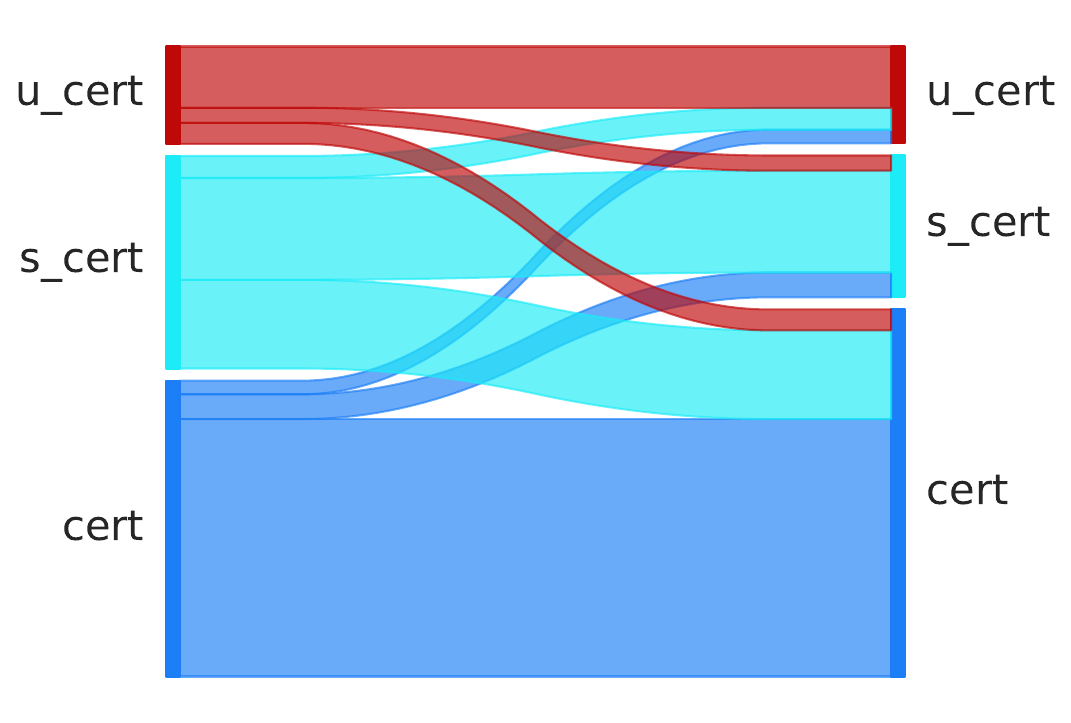}
        \caption{Science news}
        \label{fig:large-date_certain-news}
    \end{subfigure}%
    \begin{subfigure}{0.5\columnwidth}
        \centering
        \includegraphics[width=\linewidth]{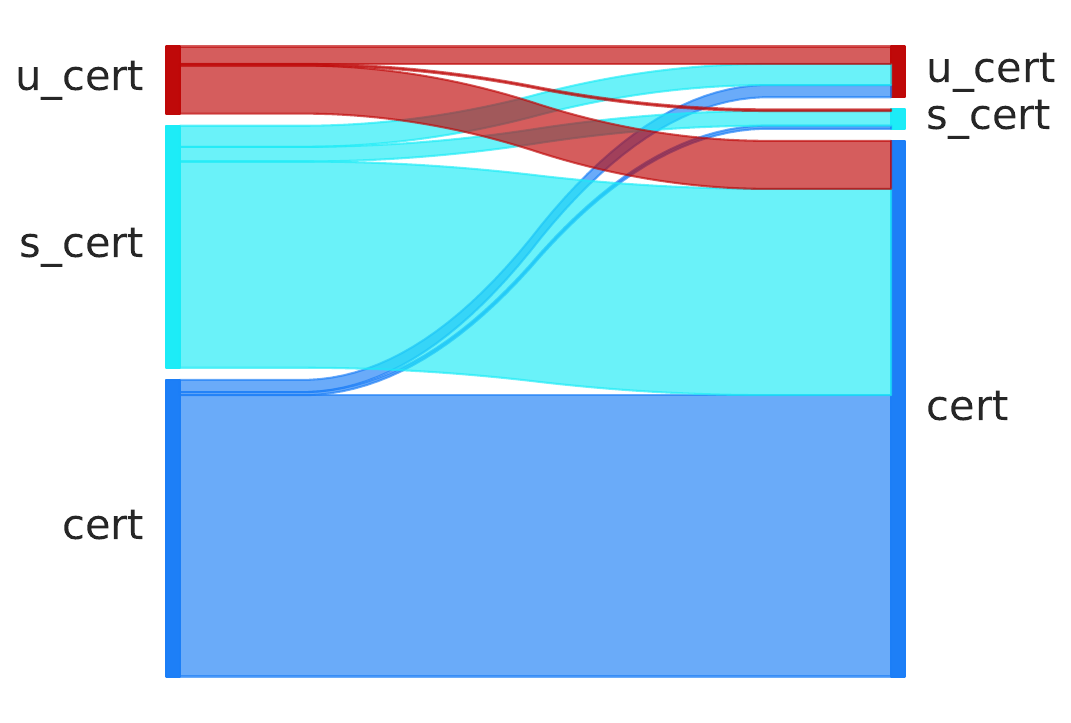}
        \caption{Tweets}
        \label{fig:large-date_certain-tweets}
    \end{subfigure}
    \caption{Changes in certainty across paired findings from science news (35,150) and  tweets (72,032).}
    \label{fig:large-data_certainty}
\end{figure}

Visualizations of changes in causality (Fig.~\ref{fig:large-data_causal}) and certainty (Fig.~\ref{fig:large-data_certainty}) across paired findings from science news (35,150) and  tweets (72,032) as Sankey diagrams. 

\subsubsection{Validating the Results from the Large-scale Analysis}
\label{appendix:validation-annotation}
For annotating the subset of the unlabeled data, we sample 108 instances across all four scientific disciplines and collect three sets of labels for each instance. One set is annotated by one of the authors, the other two sets are obtained from independent annotators who we provide with the same annotation instructions that we used to instruct the crowdworkers.

\begin{figure}
        \centering
        \includegraphics[width=0.8\linewidth]{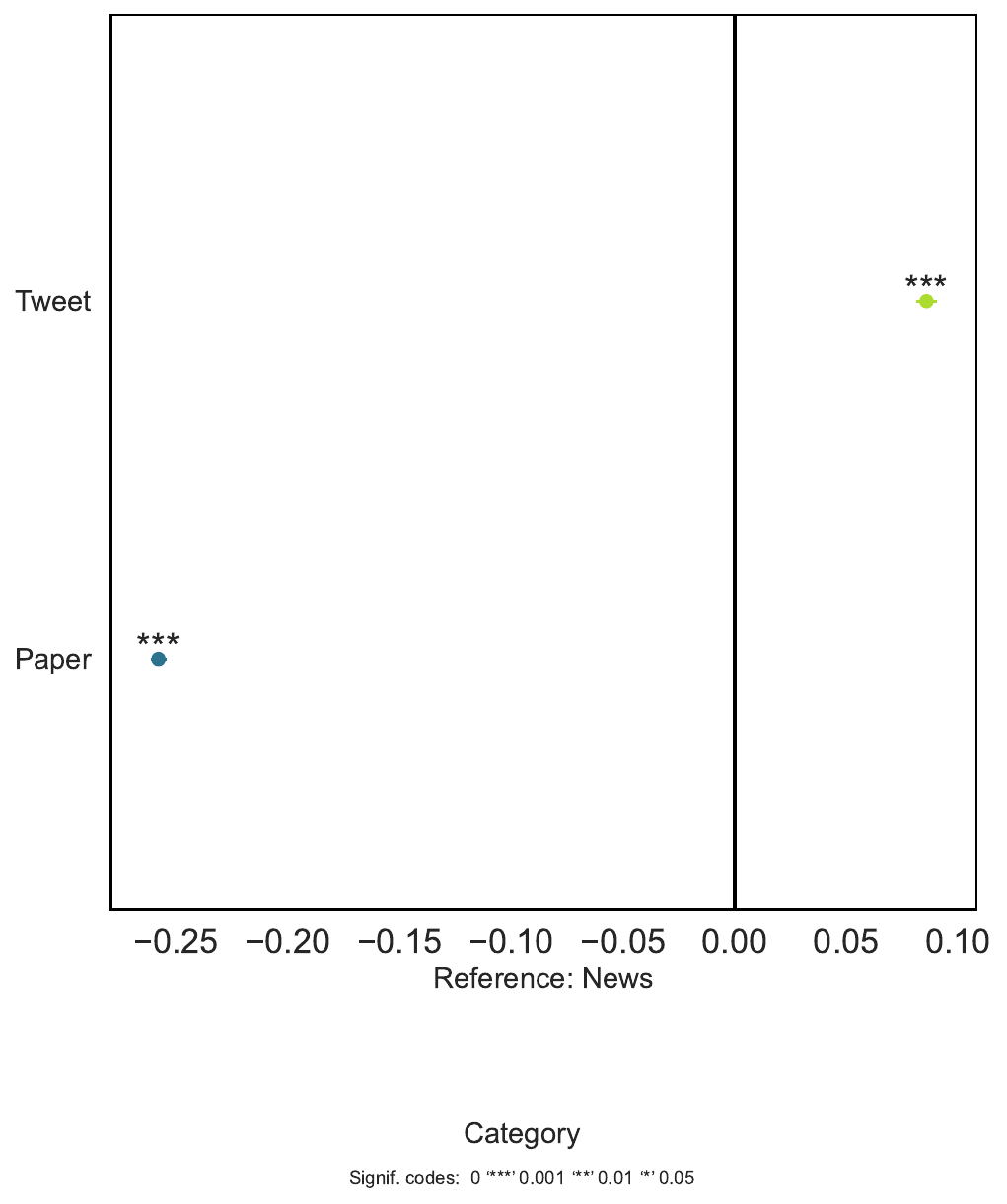}
    \caption{Plot of regression coefficients accompanying Fig. \ref{fig:large-scale-sensational_density} for predicting sensationalism score based on findings source (Paper, News, or Tweet). We see a large statistically significant effect, indicating that different sources are potentially perceived as more or less sensational.}
    \label{fig:sensationalism-significance}
\end{figure}

\end{document}